\RequirePackage{fix-cm}
\documentclass[smallcondensed]{svjour3}     
\smartqed  
\usepackage{geometry} 
\usepackage{graphicx} 
\usepackage{booktabs} 
\usepackage{array} 
\usepackage{paralist} 
\usepackage{verbatim} 
\usepackage{subfig} 
\usepackage{collcell}
\usepackage{pgf}
\usepackage{multirow}
\usepackage{supertabular}
\usepackage{longtable}
\usepackage{multicol}
\usepackage{relsize}
\usepackage{amsmath}
\usepackage{url}
\usepackage{mathptmx}      
\usepackage{natbib}

\begin{document}

\title{A Machine Learning Approach to Predicting the Smoothed Complexity of Sorting Algorithms}



\author{Bichen Shi         \and
        Michel Schellekens \and
        Georgiana Ifrim 
}


\institute{Bichen Shi \at
              Insight Centre for Data Analytics,
              University College Dublin, Belfield,
              Dublin 4, Ireland \\
               \email{bichen.shi@insight-centre.org}           
           \and
           Michel Schellekens \at
           Centre for Efficiency Oriented Languages(CEOL),
           University College Cork, Western Road,
           Cork, Ireland \\
           \email{m.schellekens@cs.ucc.ie}
          \and
          Georgiana Ifrim \at
          Insight Centre for Data Analytics,
          University College Dublin, Belfield,
           Dublin 4, Ireland \\
           \email{georgiana.ifrim@insight-centre.org}             
}

\date{Received: date / Accepted: date}

\maketitle

\begin{abstract}
Smoothed analysis is a framework for analyzing the complexity of
an algorithm, acting as a bridge between average and worst-case behaviour. 
For example, Quicksort and the Simplex algorithm are widely used in practical applications, 
despite their heavy worst-case complexity. Smoothed complexity aims to better characterize such algorithms.
Existing theoretical bounds for the smoothed complexity of sorting algorithms 
are still quite weak. Furthermore, empirically computing the smoothed complexity via its original definition is computationally infeasible, 
even for modest input sizes. In this paper, we focus on accurately 
predicting the smoothed complexity of sorting algorithms, using machine learning techniques.
We propose two regression models that take into account various properties of sorting algorithms 
and some of the known theoretical results in smoothed analysis to improve prediction quality. 
We show experimental results for predicting the smoothed complexity of Quicksort, Mergesort, and optimized Bubblesort 
for large input sizes, therefore filling the gap between known theoretical and empirical results.
\end{abstract}

\section{Introduction}
\label{sec:intro}

 {\bf S}moothed {\bf C}omplexity ({\bf SC}) was first introduced in \citep{spielman2001smoothed}, aiming to provide a more realistic view of the practical performance of algorithms compared to worst-case or average-case analysis. Motivated by the observation that, in practice, input parameters are often subject to a small degree of random noise, SC measures the expected performance of algorithms under slight random perturbations of the worst-case inputs \citep{spielman2006smoothed}. 
When worst-case is extremely rare in practice, a worst-case view can be problematic, especially for algorithms that have poor worst-case, but good average-case complexity. Average-case analysis is an important complement to worst-case analysis, providing a more comprehensive view of the problem. Nevertheless, SC, a hybrid of worst-case and average-case, provides an alternative measurement for a given algorithm \citep{spielman2002smoothed}. In practice, it is useful to understand how quickly the SC switches from worst to average case. By analyzing the worst-case inputs under perturbations, the SC intuitively indicates the probability of an algorithm encountering worst-case in practice. If the SC of an algorithm is low, then it is unlikely the algorithm will take a long time to solve practical instances, even if it has a poor worst-case complexity. 

Although very useful, SC is not easy to estimate, neither theoretically nor empirically.
The SC bounds of many algorithms have been given, including the Simplex Algorithm, that has exponential worst-case complexity but polynomial SC \citep{spielman2001smoothed,deshpande2005improved}, Quasi-Concave Minimization, an NP-hard problem, but with polynomial SC under certain conditions \citep*{spielman2009smoothed}, and Quicksort, with worst-case complexity $O(N^2)$, but SC of $O(\frac{N}{p}\ln N)$, where $p \in [0,1]$ \citep{banderier2003smoothed}. 
However, theoretical approaches to bound the SC generally require very complex proofs, and the resulting bounds are often weak. 
For sorting algorithms, \citep{Schellekens2013Modular} has shown that the gap between the exact (empirical) value of Quicksort's SC and its known bound is significant.

Recently, modular smoothed analysis \citep{schellekens2008modular} has been introduced to better estimate the value of SC for discrete cases. In \citep{schellekens2008modular} it was shown that for an algorithm that satisfies random bag preservation, its SC value can be calculated through a recurrence equation.
Although more accurate than the theoretical bounds, modular smoothed analysis currently works only for Quicksort and its median-of-three variant (M3Quicksort)  \citep{Schellekens2013Modular,Schellekens2013ModularM3Q}. Furthermore, because of its recurrence structure, the maximum input size for which it is feasible to compute the SC of Quicksort is 3000, and for M3Quicksort it is 130.

\begin{figure}[tbp]
\begin{center}
\includegraphics[trim = 0cm 0cm 0cm 0cm, clip,scale=0.3]{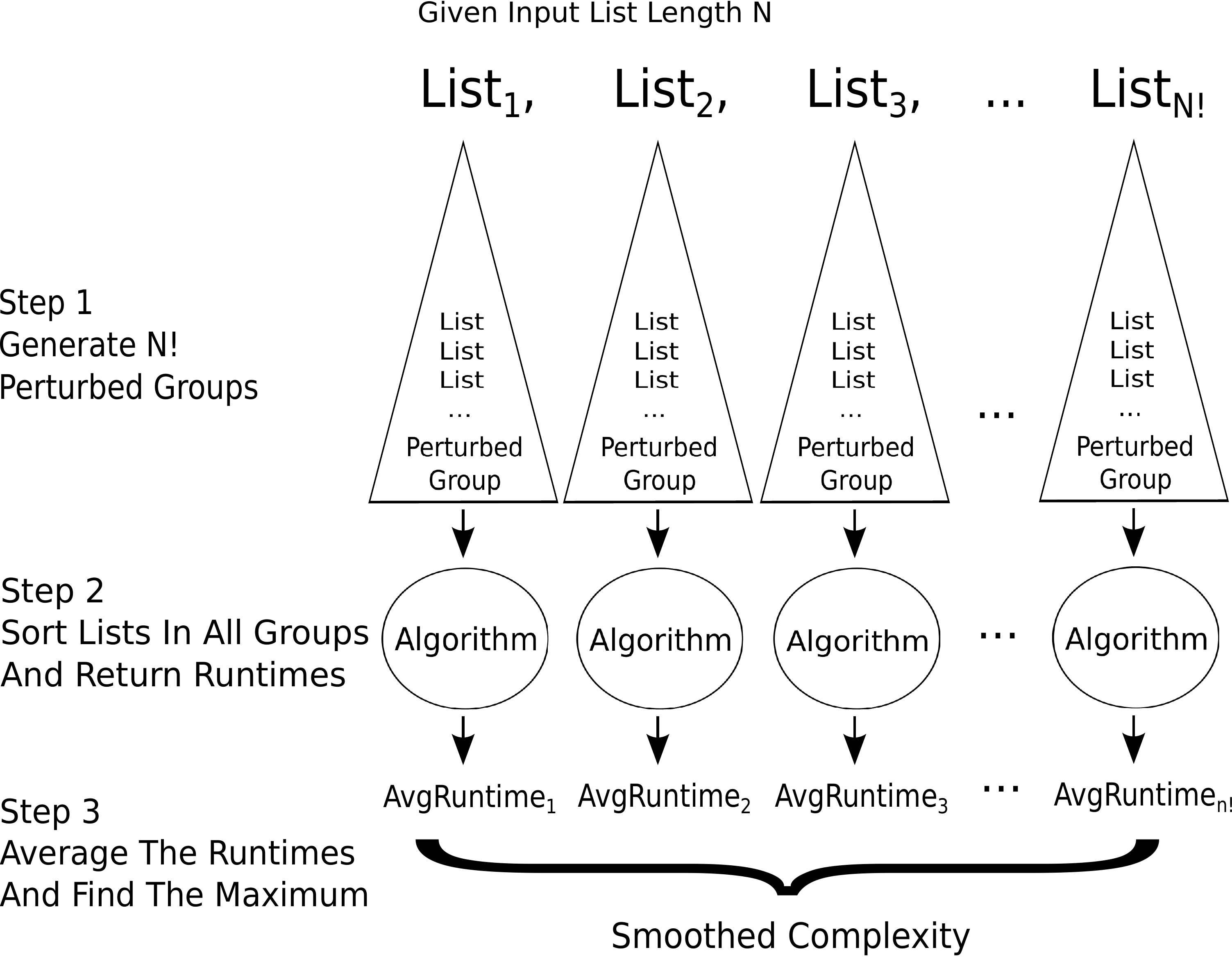}
\caption{Steps for computing the SC (maximum average runtime) of sorting algorithms empirically.}
\label{SCProcess}
\end{center}
\end{figure}

An alternative to the above approaches, is to try to compute the SC directly, using its original definition. For discrete cases, the SC under partial permutation perturbations is the maximum average runtime over all perturbed inputs \citep{spielman2006smoothed,Schellekens2013Modular}. However, the perturbation step leads to a very heavy computing process. As shown in Figure \ref{SCProcess}, to calculate the SC of an input list with length N, under partial permutations given perturbation parameter K (the degree of perturbation), we need to generate a perturbed group for each input list, take the average runtime over each perturbed group, and take the maximum of these. The complexity for empirically computing the SC of Quicksort is then
\begin{equation}\label{Complexiywithoutstore1}
O(\frac{(N!)^2 N\log (N)}{(N-K)!})
\end{equation}
quickly becoming infeasible even for very small inputs (i.e., size $N$ of list to be sorted).

{\bf Our Contribution.}
In this paper, we show how using machine learning techniques to \emph{predict} the value of SC for sorting algorithms overcomes the difficulties raised by either a theoretical or an empirical approach. This is a new point of view, since apart from theoretical bounds and some empirical results for small input sizes, there is very little information on how the SC of sorting algorithms behaves exactly. 
We formulate the SC prediction as a regression problem, and present two successful predictive models. 
Model {\bf TLR-SC} (Transformed Linear Regression for SC) turns the non-linear relationship of our selected features into a linear one, and delivers
good prediction by simply using linear regression. Model {\bf NLR-SC} (Non-linear Regression for SC) turns
a surface fitting problem into multiple curve fitting problems, and
by predicting the smoothed complexity curve by curve, gradually
predicts the entire surface. Because NLR-SC takes advantage of the
theory of smoothed analysis, it delivers good results
with very few training examples. The initial learning models are built
for Quicksort, but easily adapted to other sorting algorithms, e.g., M3Quicksort,
optimized Bubblesort and Mergesort. Previously, there were no known results on 
the SC of the latter three sorting algorithms, for large input sizes.
We believe the results in this work are useful for characterizing the
behavior of sorting algorithms in practice, and general enough so that
they could also be adapted for other interesting algorithms.

Many machine learning algorithms have been analyzed by smoothed analysis, such as k-means clustering \citep{arthur2009k} and Support Vector Machines \citep{Blum:2002:SAP:545381.545499,spielman2009smoothed}, however, to the best of our knowledge, there is no previous research on using machine learning algorithms to predict the SC of sorting algorithms. 
Due to recent modular smoothed analysis results, we can compute the SC of Quicksort exactly, for medium size inputs (e.g., $N \leq 3000$). For other sorting algorithms, currently the SC can only be computed using its original definition, and thus, due to computational requirements, only for small input sizes (e.g., $N \leq 100$), limiting our understanding of the general SC behaviour. This also means that we are faced with a lack of ground truth data to train machine learning algorithms for predicting the SC. To this end, in this work, we use a combination of modular smoothed analysis and empirical approaches to generate ground truth data.
We formulate predicting the numeric value of SC of sorting algorithms as a regression problem. The techniques to handle regression, curve fitting and surface fitting problems in the machine learning area are quite mature \citep{hastie:slbook13}. The gist of this work is how to gather ground truth data, identify good features and build appropriate learning models, so that by training on behaviour data of sorting algorithms on small inputs (where it is relatively easy to gather ground truth data), we can accurately predict the SC of the sorting algorithm for large inputs.

\section{Discrete Smoothed Complexity}
\label{sec:dsc}

While worst-case complexity refers to the maximum running time of an algorithm acting on every input, the SC is a smoothed version of worst-case complexity, that considers the maximum average running time of an algorithm acting on the perturbations of every input. The degree of perturbation is measured by a parameter $\sigma.$ SC can be explained as a function of $\sigma$ which interpolates between the worst-case and average-case running times. When $\sigma$ goes to 0, then the SC is equal to the worst-case complexity; whereas, if $\sigma$ goes to 1, then the SC becomes the average-case complexity, and in practice, it is useful to understand how quickly the SC switches from worst to average case. The quicker the SC switches, the more unlikely the worst-case appeares in practice.

SC was originally defined for continuous cases using Gaussian perturbations  \citep{spielman2001smoothed}. The discrete version of SC was introduced in \citep{banderier2003smoothed} and extended in \citep{Schellekens2013Modular}. 
In this work, we use the partial permutation perturbation definition of \citep{Schellekens2013Modular}, where $\sigma = \frac{K}{N} \; (0 \leq K\leq N)$, N is the length of the input list:
\begin{definition} 
\label{define-partial-permutation}
A $\sigma$-partial permutation of S is a random sequence $S' = (S_1', S_2', \dots S_N')$ obtained from $S = (S_1, S_2, \dots S_N)$ in two steps.
\begin{enumerate}
\item $K$ elements of $S$ are selected at random.
\item Choose one of the $K!$ permutations of these elements (uniformly at random) and rearrange them in that order, leaving the positions of all the other elements fixed.
\end{enumerate}
\end{definition}

Table \ref{Partial-Permutations} shows the partial permutations for $N = 3$ on the set of permutations of $\{1,2,3\}.$ When $K=0$, the list is not perturbed. When $K=1$, the perturbed group contains the list itself, repeated $N$ times. When $K=N$, the perturbed group is the set of permutations of size $N$, $\sum_N$.
\begin{table}[htbp]
\caption{The partial permutations for $N = 3$ on the set of permutations of $\{1,2,3\}.$}
\label{Partial-Permutations}
\centerline{
\begin{tabular}{|c|c|c|c|c|c}
\hline
K=0 & K=1 & K=2 & K=3\\
\hline
123 & 123, 123, 123 & 123, 123, 123, 132, 213, 321& 123, 132, 213, 231, 312, 321\\
\hline
132 & 132, 132, 132 & 132, 132, 132, 123, 312, 231& 132, 123, 312, 321, 213, 231,\\
\hline
213 & 213, 213, 213 & 213, 213, 213, 231, 123, 312 & 213, 231, 123, 132, 312, 321\\
\hline
231 & 231, 231, 231 & 231, 231, 231, 213, 321, 132 & 231, 213, 321, 312, 123, 132\\
\hline
312 & 312, 312, 312 & 312, 312, 312, 321, 132, 213 & 312, 321, 132, 123, 213, 231\\
\hline
321 & 321, 321, 321 & 321, 321, 321, 312, 231, 123 & 321, 312, 231, 213, 123, 132 \\
\hline
\end{tabular}
}
\end{table}

Having the partial permutation perturbation definition, we now give the definition of the discrete SC.

\begin{definition} \citep{Schellekens2013Modular}\label{Definition-SC}
 Given a problem P with input sequence domain $\sum_N$, let A be an algorithm for solving P. Let $\overline{T}_A(\mathcal{I})$ be the average running time $\overline{T}$ of an algorithm $A$ on an input collection $\mathcal{I} \subset \sum_N$. The SC, $T_{A}^S (N,K)$, of the algorithm $A$ is defined by:
\begin{equation}
T_{A}^S (N,K) = max_{S \in \sum_N}(\overline{T}_A (Pert_{K,N}(S)))
\end{equation} where $Pert_{K,N}(S)$ is the perturbed group of S under partial permutations, the degree of which is defined by $K.$
\end{definition}

In this work, the algorithms are comparison-based, therefore $T_A(x)$ is measured as the number of comparisons $A$ performs when computing the output on input $x$.

\subsection{Modular Smoothed Analysis}
\label{ssec:msa}
Modular smoothed analysis was recently introduced in  \citep{Schellekens2013Modular,Schellekens2013ModularM3Q}. It is a simplification of traditional SC analysis by smoothing out the perturbations over the computation. For algorithms that are random bag preserving \citep{schellekens2008modular}, the number of comparisons of the algorithm running on an input can be captured and calculated through a recurrence equation. The two equations for the modular SC of Quicksort and M3Quicksort are shown below (Equations~\ref{msaQ},\ref{msaMQ}). Using these two formulas we collect ground truth data for Quicksort and its median-of-three variant for training and evaluating our supervised learning method.
The modular recurrence equation for the SC of Quicksort, $f(N,K),$ is \citep{Schellekens2013Modular}:
\begin{equation}
f(N,K) =  (N-1) + \sum_{j=1}^{N} \beta_{N+1-j}^{N} f(j-1,K) + \sum_{j=1}^{N} \beta_j^{N} f(j-1,K),
\label{msaQ}
\end{equation}
\noindent where
$$\beta_{N}^{N} = \frac{N - K +  1}{N},$$
\noindent and
$$\beta_{j}^{N}[j\ne N] = \frac{(K-1)}{N(N-1)}.$$\\

\noindent The recurrence equation for the SC of M3Quicksort is  \citep{Schellekens2013ModularM3Q}:\\

\begin{equation}
f(N,K) =  (N-1) + 1 + \sum_{j=2}^{N-1} \beta_{N+1-j}^{N-1} f(j-1,K) + \sum_{j=2}^{N-1} \beta_j^{N-1} f(j-1,K)
\label{msaMQ}
\end{equation}
\noindent where
$$\beta_j^{N-1} = \frac{2!(j-1)}{K(N-2)\binom{N}{K}} \Bigg ( 2\binom{N-4}{K-2}  + 2\frac{(K+1)}{(K-1)}\binom{N-4}{K-3} + \frac{3(N-j)}{(K-1)}\binom{N-4}{K-4} \Bigg ),\ \\ \quad 2 \le j \le N-2,$$
\noindent and
$$\beta_{N-1}^{N-1} = \Bigg \{ K!\binom{N-3}{K}+ (K-1)!\binom{N-3}{K-1}( 2 + K ) + 2!( K-2)!\binom{N-3}{K-2}(2K-1)$$
$$ + 3!(K-2)!\binom{N-3}{K-3} \Bigg \}  \frac{(N-K)!}{N!}$$

Modular SC values are closer to the traditional SC ones, as compared to the existing mathematical bounds \citep{Schellekens2013Modular}. 

\section{Sorting Algorithms}
This paper focuses on analysing and predicting the SC of four sorting algorithms: Quicksort, M3Quicksort, optimized Bubblesort and Mergesort. Their worst-case, average-case and smoothed complexity are listed in Table \ref{table:complexity}.

\begin{table}[thbp]
\caption{The worst-case, average-case, and smoothed complexity (if known) of Quicksort, M3Quicksort, optimized Bubblesort and Mergesort.}
\label{table:complexity}
\centerline{
\begin{tabular}{|c|c|c|c|}\hline\hline
Sorting Algorithm		& Worst-Case	& Average-Case 	& Smoothed Complexity \\\hline
Quicksort 			& $O(N^2)$ 	& $O(NlogN)$	& $O(\frac{N}{p} \log(N))$\\
M3Quicksort 	& $O(N^2)$ 	& $O(NlogN)$	& NA \\
Optimized Bubblesort 		& $O(N^2)$ 	& $O(N^2)$	& NA \\
Mergesort 			& $O(NlogN)$ 	& $O(NlogN)$	& NA \\ \hline\hline
\end{tabular}
}

\end{table}

M3Quicksort is a variation of Quicksort. The classical version of Quicksort selects the first element of the list as a pivot, while M3Quicksort first compares the first, the median and the last element, and selects as pivot the element whose value is the median of the three. By doing so, M3Quicksort is 30\% - 50\% faster than the original algorithm.

Bubblesort is simple to implement, and it is also easy to track its comparisons. The normal Bubblesort has a constant runtime for all inputs, and it is not desirable as a testing algorithm for the SC, which analyzes the transition between the worst-case and the average-case complexity. Therefore, we focus here on optimized Bubblesort \citep{DBLP:books/aw/Knuth73}. A tracker is added in Bubblesort to show whether or not elements were swapped, so the algorithm can stop running earlier if the list has been sorted. As a result, optimized Bubblesort works faster on part of the inputs and its average-case runtime will be smaller than its worst-case runtime, although its two complexities are on the same scale.

Mergesort is the standard version. Note that although its worst-case and average-case complexities are on the same scale, its worst-case and average-case {\it runtimes} do not have the same value.

\section{Data Collection}
\label{sec:datacollection}

In this section we describe our process of collecting ground truth data for learning prediction models. All our data and code is available upon request for research purposes.
\subsection{Empirical Approach}
Although \citep{banderier2003smoothed} has proven that the SC of Quicksort is $O(\frac{N}{p} \ln(N))$, this bound is not accurate enough (see \citep{Schellekens2013Modular} for details) to allow us training a supervised machine learning approach. To obtain better ground truth data for the SC, we first use an experimental approach to compute the SC exactly, by following definitions in \citep{Schellekens2013Modular}.
The steps for calculating the SC of a sorting algorithm \emph{A} for given input lists length N are (Figure \ref{SCProcess}):\\
\textbf{Step1}. Generate a perturbed group for each input list, under the partial permutation perturbation. Given the perturbation parameter K, and following Definition \ref{Partial-Permutations}, the size of each perturbed group is
\begin{equation}\label{sizeofPbag}
\binom{N}{K}K! = \frac{N!}{K!(N-K)!}K! 
= \frac{N!}{(N-K)!}
\end{equation}
The total number of permutations in all perturbed groups is equal to the number of the input lists, which is $N!$, multiplied with the size of each perturbed group:
\begin{equation}\label{sizeofPermutations}
\frac{(N!)^2}{(N-K)!}
\end{equation}
\textbf{Step2}. For all permutations in every perturbed groups, compute their runtimes\footnote{Runtime is the number of comparisons in our case, as we only consider sorting algorithms in this work.} under sorting algorithm \emph{A}. Denote the average-case complexity of \emph{A} as $A(N)$, then the complexity of this process is
\begin{equation}\label{Complexiywithoutstore}
O(\frac{(N!)^2\cdot A(N)}{(N-K)!})
\end{equation}
\textbf{Step3}. Calculate the average runtime for each perturbed group, then select the maximum average runtime among all perturbed groups, i.e., the SC. Compared to Equation \ref{Complexiywithoutstore}, the complexity of calculating the average and the maximum is too low to be considered. Therefore the complexity of computing SC of \emph{A} is the same as Equation \ref{Complexiywithoutstore}.\\

\begin{figure}[pt]
\centering
\subfloat[Subfigure 1 list of figures text][Quicksort]{
\includegraphics[width=0.55\textwidth]{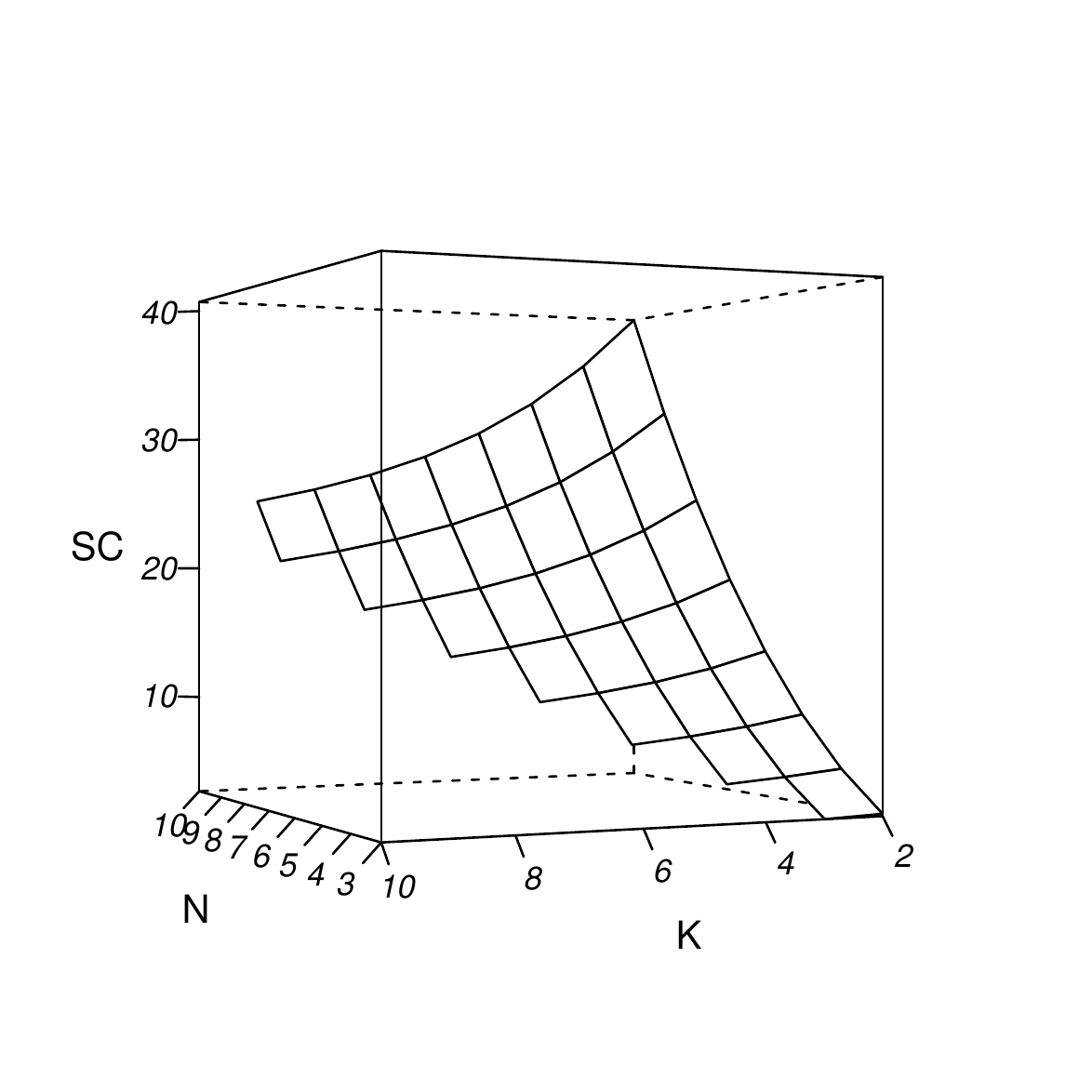}
\label{3D-ExperimentalSC}}
\subfloat[Subfigure 2 list of figures text][M3Quicksort]{
\includegraphics[width=0.55\textwidth]{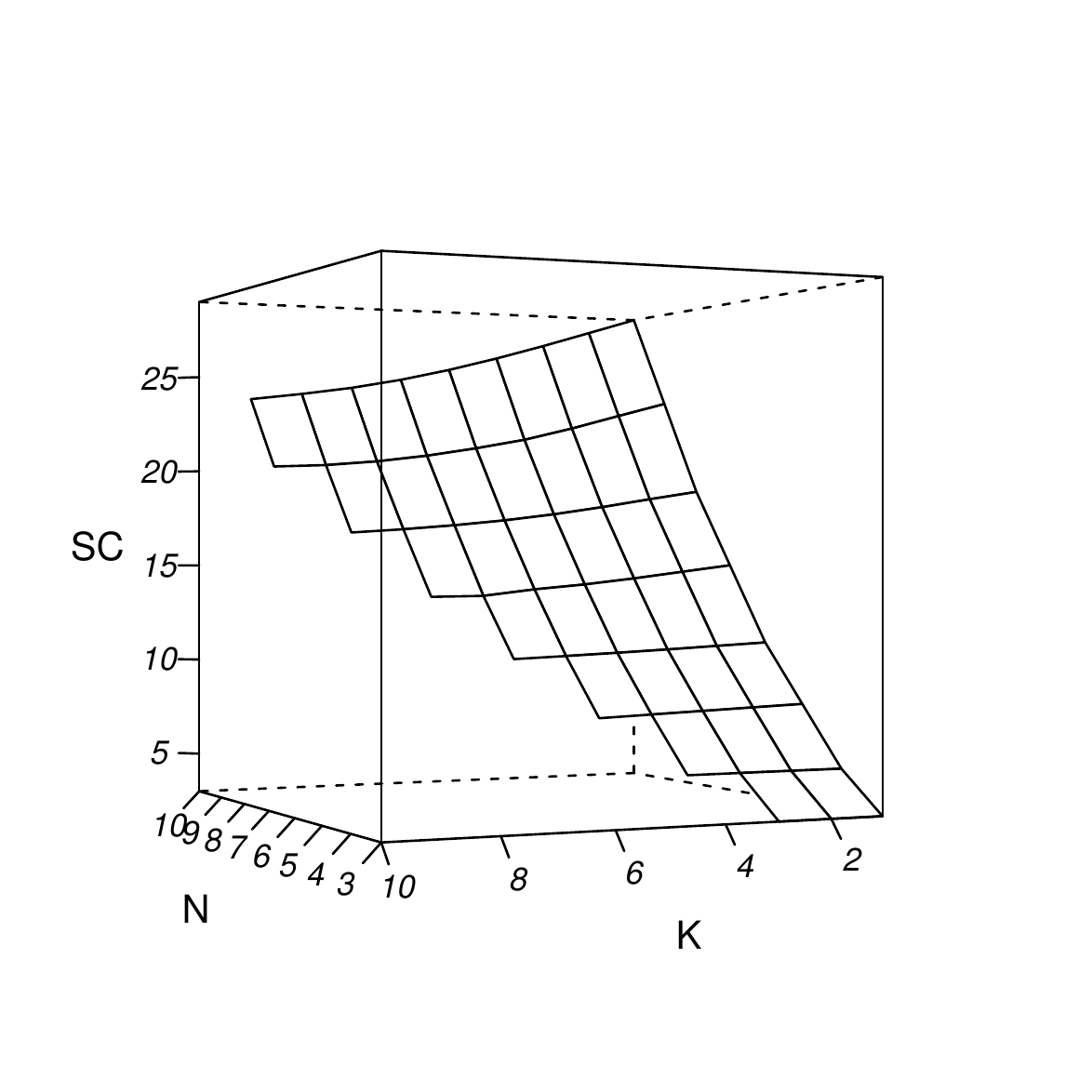}
\label{3D-ExperimentalMOT}}
\\
\subfloat[Subfigure 3 list of figures text][Bubblesort]{
\includegraphics[width=0.55\textwidth]{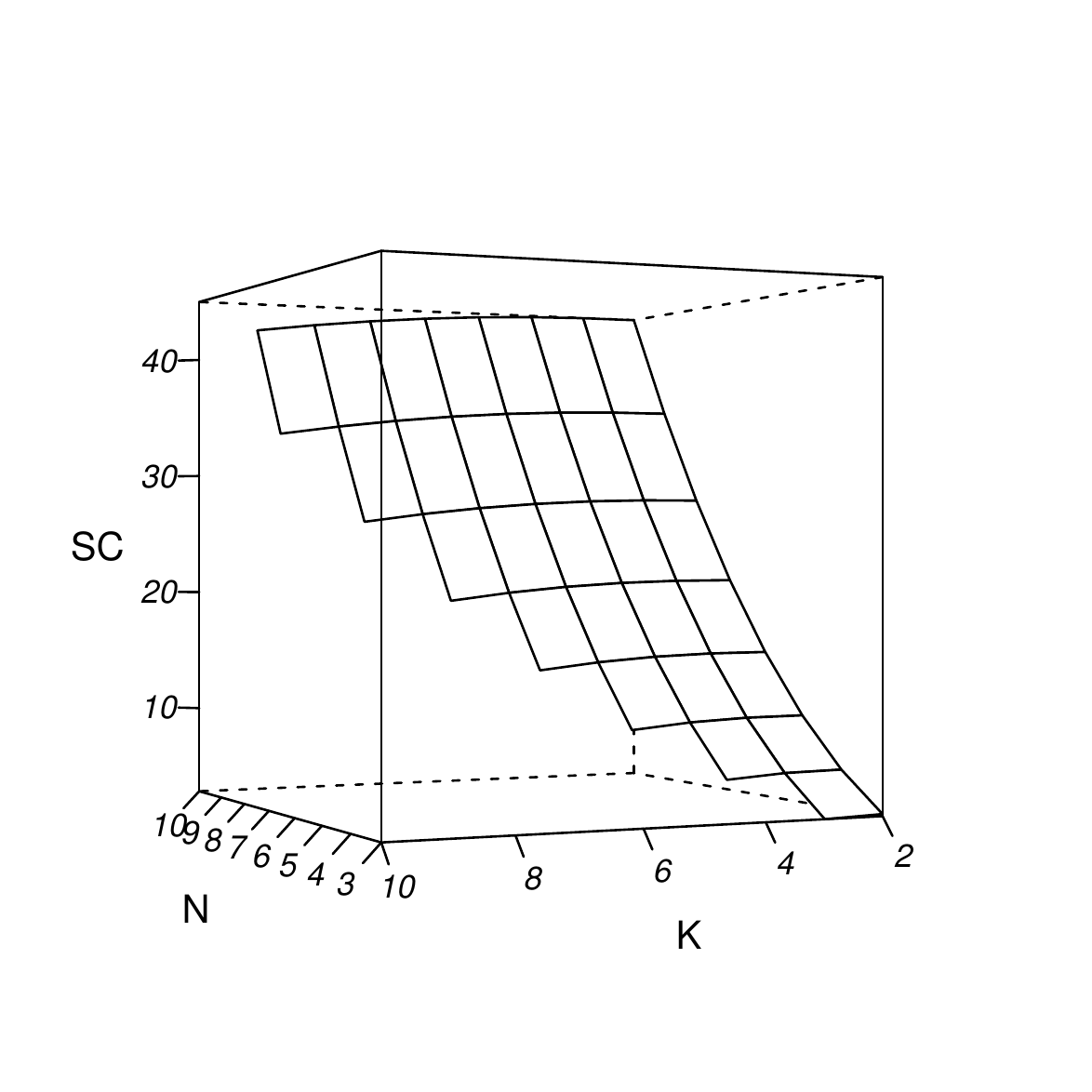}
\label{3D-ExperimentalBS}}
\subfloat[Subfigure 4 list of figures text][Mergesort]{
\includegraphics[width=0.55\textwidth]{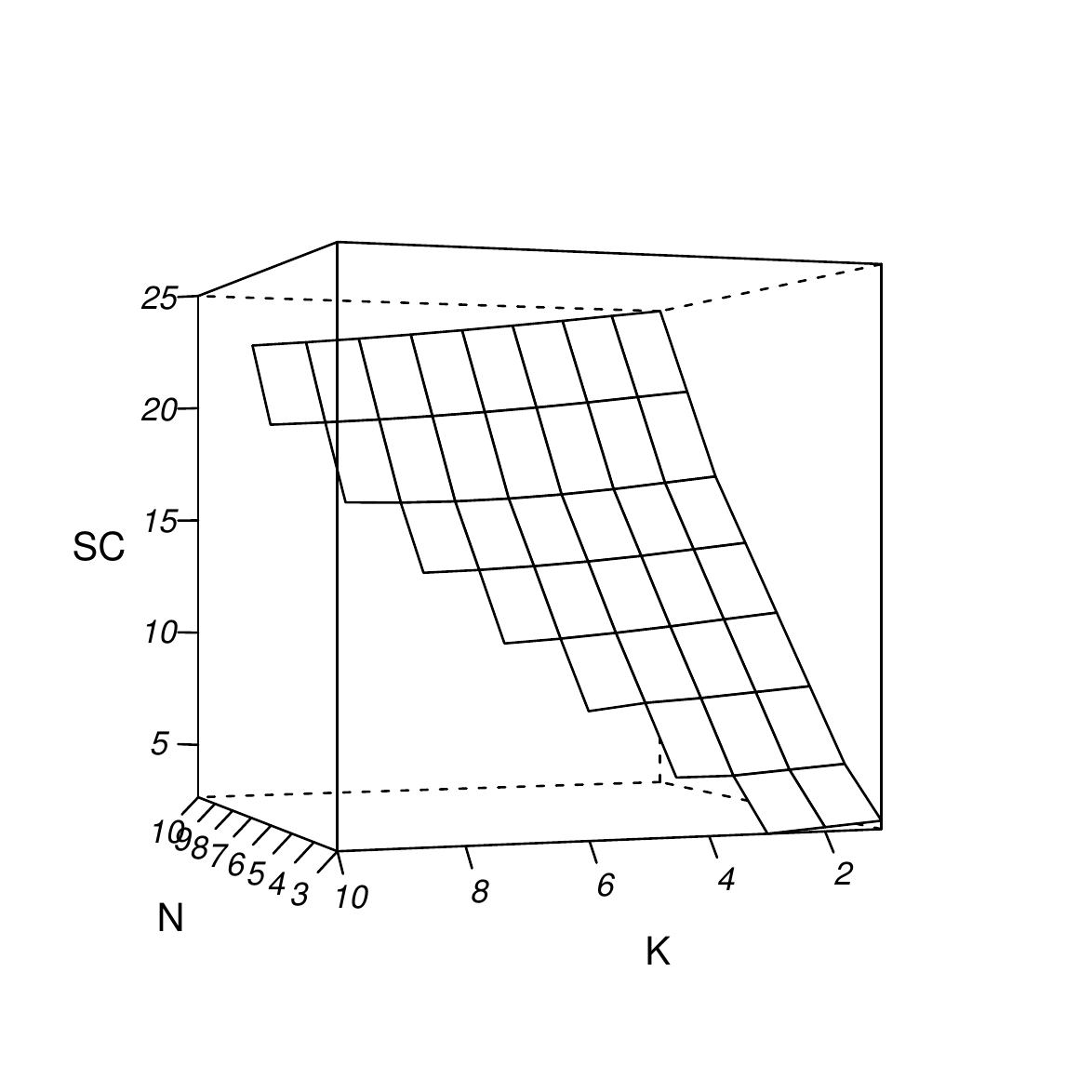}
\label{3D-ExperimentalMS}}
\caption{The Smoothed Complexity of four sorting algorithms calculated by the experimental approach, where $N \leq 10,  2\leq K \leq N$.}
\label{All-3D-ExperimentalSC}
\end{figure}

We can see how computationally heavy this process is. Even though we can store the runtime of the $N!$ permutations into memory to avoid repetitive computation, the complexity of computing the SC is still 
\begin{equation}\label{Complexiywithstore}
O(N!\cdot A(N))
\end{equation}
For Quicksort, the complexity of computing the SC is $O(N!N\log (N))$, while for Bubblesort, the complexity is $O(N!N^2)$. However, since empirical data is essential as a ground truth for a machine learning approach, we generate some of our data with this approach.
We compute the SC of Quicksort, M3Quicksort, optimized Bubblesort, and Mergesort. Our code uses hill climbing and Quicksort-specific worst-case permutation results \citep{claire:mst}, to push the input size for which we can empirically obtain SC values.
Plots of the results of the four algorithms are shown in Figure \ref{All-3D-ExperimentalSC}, for list length $N=10, \; 2 \leq K \leq N$. 

{\bf Limitations of the Empirical Approach:} 
Through the experimental approach, we can obtain the SC of sorting algorithms for very limited N and K, due to the computational burden. For $K=2$, we can compute the SC for Quicksort only up to $N=200$, and for $K=N$, only up to $N=10$. For algorithms that are less efficient than Quicksort, the data we can collect this way is much less.
Such a dataset alone is too small to be useful for training and testing our learning system. Therefore, in the next section, we show how to use modular analysis to gather more data.

\subsection{Modular Smoothed Analysis}

Modular smoothed analysis \citep{Schellekens2013Modular} provides another way to calculate the SC for Quicksort and its median-of-three variant. More precisely, it gives recursive formulas, that are parameterized by the list length N and perturbation parameter K.
Using the same amount of time as the empirical approach, we can collect a thousand times more data through the modular approach. For Quicksort, the maximum list length N for which the SC can be computed is 3000. 
For M3Quicksort, we can collect the SC for N up to 130.
Unfortunately, when $N > 130$, the factorial calculation within the formula makes the computation unmanageable. In addition, the formula only works when $K \geq 4$, and the SC of $K=2, K=3$ cannot be obtained through the modular approach.

{\bf Limitations of the Modular Approach:} Due to the recursive structure of modular analysis equations, this aproach also quickly becomes infeasible with increasing input N. Additionally, for now, modular smoothed analysis results are known only for Quicksort and M3Quicksort, while we require ground truth data to validate learning models for more sorting algorithms, e.g., Bubblesort and Mergesort.

\section{Data Analysis}
\label{sec:dataanalysis}

Section \ref{sec:datacollection} discussed the collection of the ground truth data for building SC prediction algorithms, i.e., given an input list of length $N$, and a perturbation parameter $K$, we have computed the value of SC, for four sorting algorithms. 
For Quicksort and M3Quicksort, we used the modular approach, and for Bubblesort and Mergesort,  the empirical approach. Table \ref{sample data table} shows sample ground truth data for Quicksort.
\begin{table}[htbp]
\caption{Sample ground truth data for the SC of Quicksort.}
\label{sample data table}
\begin{center}
\begin{tabular}{|c|c|c|} \hline\hline
N  & K & SC \\ \hline
10 & 2	& 	39.7305\\
10 & 3	&35.6077\\
10	&4	&32.4413\\
10 & ... & ... \\
10	&10	&24.4373\\
15	&2	&91.8248\\
15	&3	&81.6957\\
15	&4	&73.8442\\
...	&...	&...\\ \hline\hline
\end{tabular}
\end{center}
\end{table}
Figure \ref{3D-ExperimentalSC} shows the relationships between $N$, $K$ and the SC of Quicksort. If the value of $N$ increases, no matter what value $K$ is, the SC increases as well. This is reasonable, since the execution time will be longer in general for sorting algorithms, when the input list length is greater. If the value of $K$ increases, no matter what value $N$ is, the SC decreases. Because the SC is the hybrid of worst-case and average-case analysis, if K increases, SC will tend from the worst-case towards the average-case behavior \citep{spielman2001smoothed}. 
From Figures \ref{3D-ExperimentalMOT}, \ref{3D-ExperimentalBS}, \ref{3D-ExperimentalMS} we can see similar patterns also exist for the SC of M3Quicksort, Bubblesort and Mergesort.

\subsection{Fixed N}

\begin{figure}[htbp]
\begin{center}
\includegraphics[width=10cm,height = 10cm]{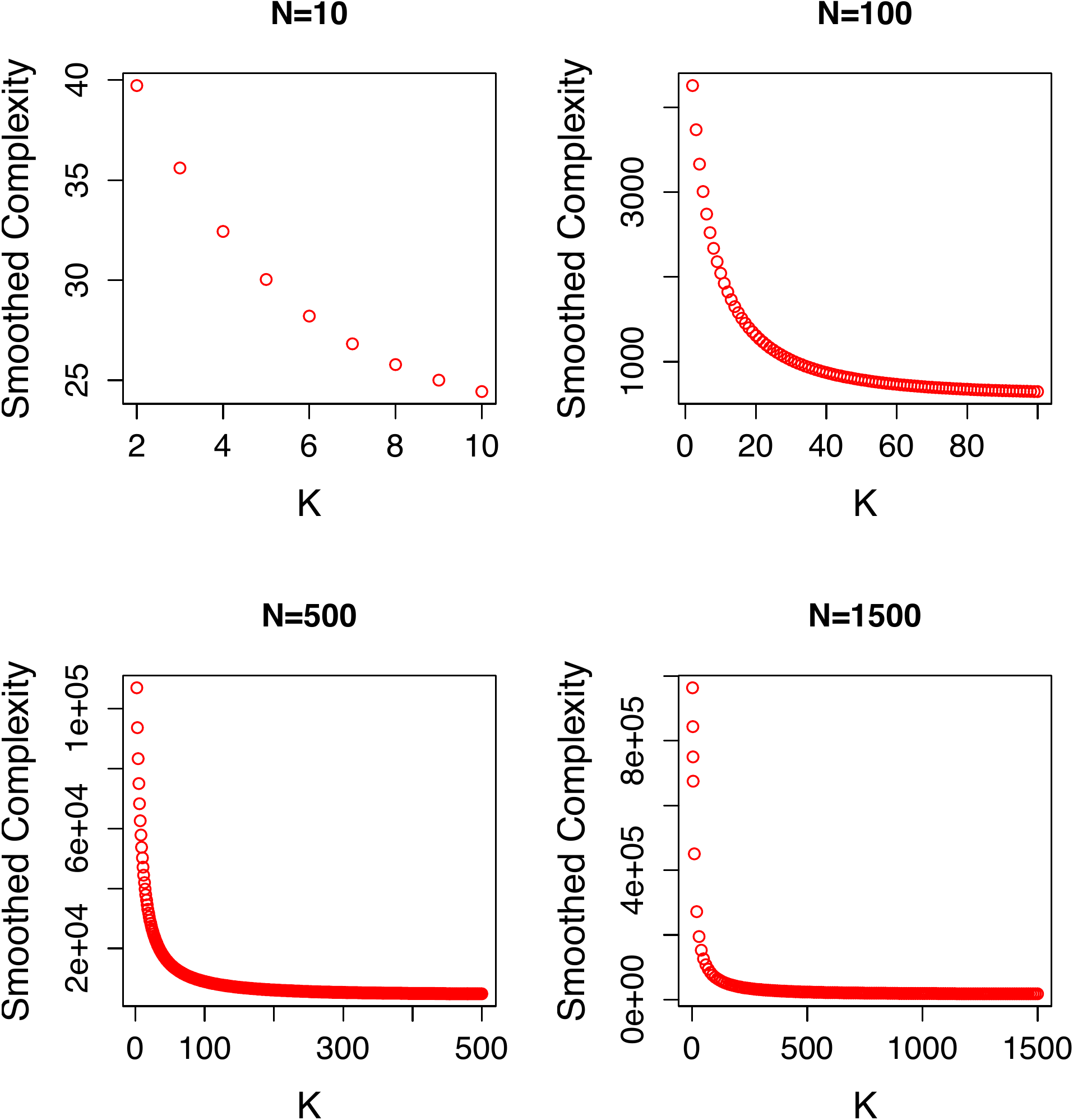}
\caption{The SC of Quicksort where $N = 10,100,500,1500$, $2\leq K \leq N$.}
\label{FixN-SC-K-100-1500}
\end{center}
\end{figure}

By fixing the value of N, we consider the relationship between the SC and K only. Figure \ref{FixN-SC-K-100-1500} shows how the SC of Quicksort decreases while K increases, for $N=10,100,500,1500$, $2\leq K \leq N$. When $K=2$, the value of the SC is very close to the worst-case complexity, and when $K=N$, the value is equal to the average-case complexity \citep{spielman2001smoothed}. Note that, the larger the value of N, the quicker the SC decreases while K increases. 

\begin{figure}[htbp]
\centerline{
\includegraphics[trim = 0cm 0cm -0.5cm 0.5cm, clip,width=15cm]{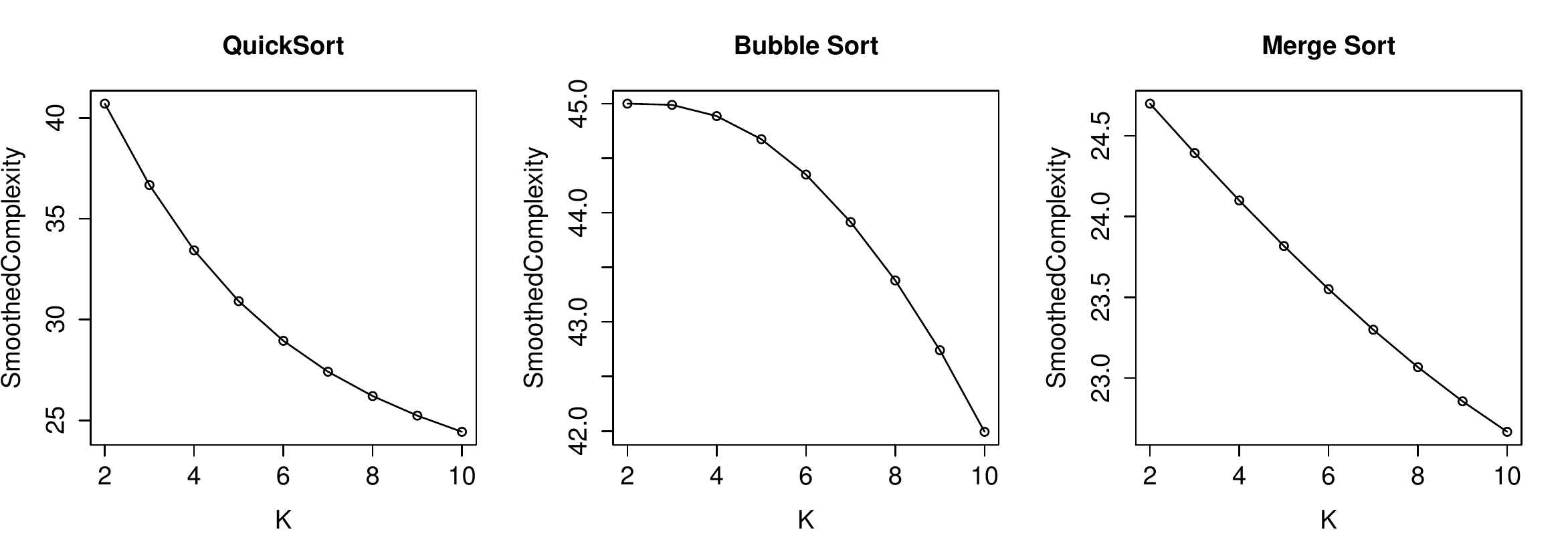}
}
\caption{Data shape of SC for varying K, for Quicksort, Bubblesort, Mergesort, $N=10$.}
\label{Fix-N-Algorithms}
\end{figure}

By looking at the Figure \ref{FixN-SC-K-100-1500} where $N=500$, we can clearly see the tipping point where SC steadily turns to average-case complexity. Such point indicates how likely the algorithm encounters a worst-case in practice, and the quicker the tipping point appears, the better the SC of the algorithm. The data shape of the SC depends on the sorting algorithm, as shown in Figure \ref{Fix-N-Algorithms}. The relationships between the SC and K of Quicksort, optimized Bubblesort and Mergesort are completely different. While K increases, the SC of Quicksort decreases quickest, the second is Mergesort, and the last is Bubblesort. This explains why QuickSort performs very well in practice although its worst-case complexity is $O(N^2)$, and also shows how the SC encodes this behaviour.

\subsection{Fixed K}
\begin{figure}[htbp]
\centerline{
\includegraphics[trim = 0cm 0.5cm 0cm 0.5cm, clip,width=9cm]{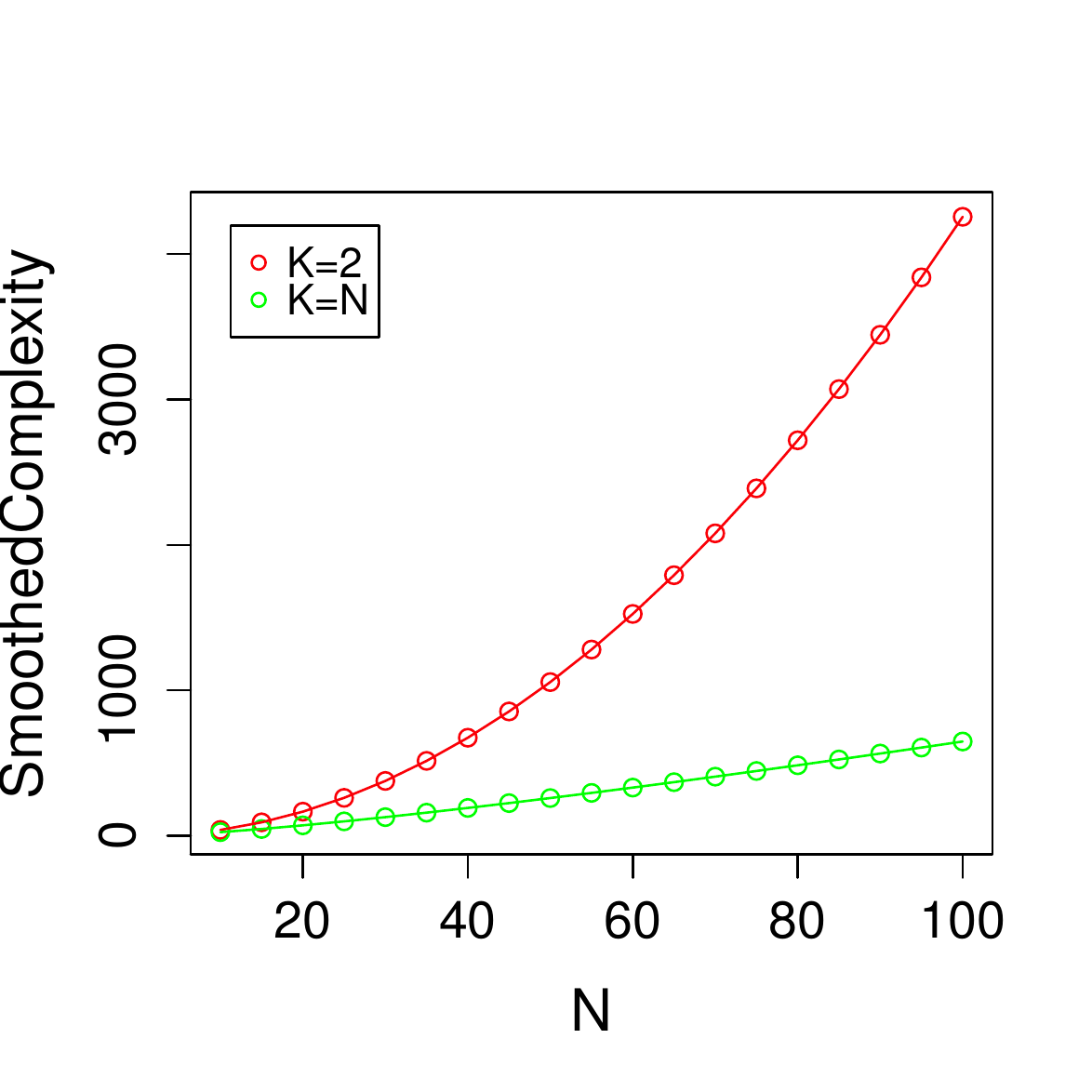}
}
\caption{The SC of Quicksort when fixing K=2 and K=N.}
\label{FixK-SC-N-10-100}
\end{figure}

With K fixed, we analyze the relationship between the SC and N. Figure \ref{FixK-SC-N-10-100} shows how the SC of Quicksort increases when $K=2$, and when $K=N$. The range of N is from 5 to 100. The value of the SC given $K=2$, links to the worst-case behavior of the algorithm, which is $O(N^2)$ for Quicksort, and similarly, given $K=N$, it links to the average-case behavior, which is $O(N\log  N)$. 

\subsection{Feature Selection}

Due to the definition of SC, when selecting features for designing our learning models, N and K are the first choice. N is the input list length and K is the perturbation parameter varying from 2 to N.
In our experiments, we also found Runtime-based features to be helpful for predicting the SC. Runtime is the comparison time of an input list in a sorting algorithm, and MaxRuntime is the maximum Runtime among all input lists, given N. The average time complexity for computing Runtime of Quicksort over all inputs is $O(N\log N)$. We define AvgRuntime as the value of average Runtime of an input's perturbed group in a sorting algorithm. Note that the SC is the maximum AvgRuntime (MaxAvgRuntime) that can be found among all inputs.
We have found that Runtime and AvgRuntime do not improve prediction, but the MaxRuntime does. 
For each N, there is only one MaxRuntime value. MaxRuntime is fairly easy to compute compared to the SC, and it can act as a scaling factor to indicate an appropriate SC value for the model.  For Quicksort we compute MaxRuntime using the worst-case permutation. For the other 3 algorithms, we use hill climbing.
Table \ref{Feature-table} shows the definition of important terms for this section.
\begin{table}[htbp]
\caption{Summary of important terms/features for our prediction models.}
\label{Feature-table}
\centerline{
\begin{tabular}{|l|m{9.6cm}|l} \hline\hline
Terms & Definition \\ \hline\hline
N & Input list length \\ \hline
K & Perturbation parameter, varies from 2 to N\\ \hline
MaxRuntime & Maximum Runtime among all input lists of size N. For each N, there is only one corresponding MaxRuntime \\ \hline
MaxAvgRuntime & Maximum AvgRuntime among all input lists of size N, given K. For each N and K combination, there is only one corresponding MaxAvgRuntime, aka the SC \\ \hline\hline
\end{tabular}
}
\end{table}

\section{Prediction Models for the SC}
\label{sec:predmodels}
In this section we present our evaluation methodology and our two approaches for predicting the SC of sorting algorithms.

\subsection{Evaluation Metrics}

We use three classical measures to evaluate the prediction quality of the models tested. Assume the size of the test set is $n$; the actual target attribute values in the test set are $a_1,a_2,\dots,a_n$; the predicted values on the test instances are $p_1,p_2,\dots,p_n$. The Mean Absolute Error (MAE) is \citep{witten2011data}:
\begin{equation}
MAE = \frac{|p_1-a_1|+\dots+|p_n-a_n|}{n}
\end{equation}

When the relative rather than the absolute error values are more important, we use Mean Absolute Percentage Error (MAPE).
\begin{equation}
MAPE = \frac{|\frac{p_1-a_1}{a_1}|+|\frac{p_2-a_2}{a_2}|+\dots+|\frac{p_n-a_n}{a_n}|}{n}\times 100\%
\end{equation}

We also show the Root Mean Squared Error (RMSE), which is more sensitive to outliers \citep{witten2011data}:
\begin{equation}
RMSE = \sqrt{\frac{(p_1-a_1)^2+\dots+(p_n-a_n)^2}{n}}
\end{equation}
We use the R language environment for analysing our data and building predictive models, since it offers more flexibility in manipulating/visualising data. All our data and R code are available upon request for research purposes.

\subsection{Model TLR-SC}
\label{ssec:tlrsc}
In order to predict the SC we have first tested several feature combinations (e.g., N, K, Runtime, MaxRuntime) and several built-in regression algorithms of the open source machine learning software Weka \citep{witten2011data}. 
Nevertheless, straightforward application of WEKA algorithms did not work well for predicting the SC, in particular, for the scenario we are interested in: training on small input sizes and predicting/testing on large input sizes. Most WEKA algorithms delivered MAPE around 20\%, when trained on data with $N \leq 20$ and tested on data with $N \geq 40$.

In this section, we discuss several modelling approaches, and propose a first model for accurately predicting the SC, named {\bf TLR-SC} (Transformed Linear Regression for Smoothed Complexity). The idea behind TLR-SC is 
to build new features that better capture the nature of the relationship between the SC and input data characteristics. For example, 
as shown in Figure  \ref{3D-ExperimentalSC}, for Quicksort, the SC has a nonlinear relationship with K and N.
\begin{figure}[tbp]
\centerline{
\includegraphics[width=9cm]{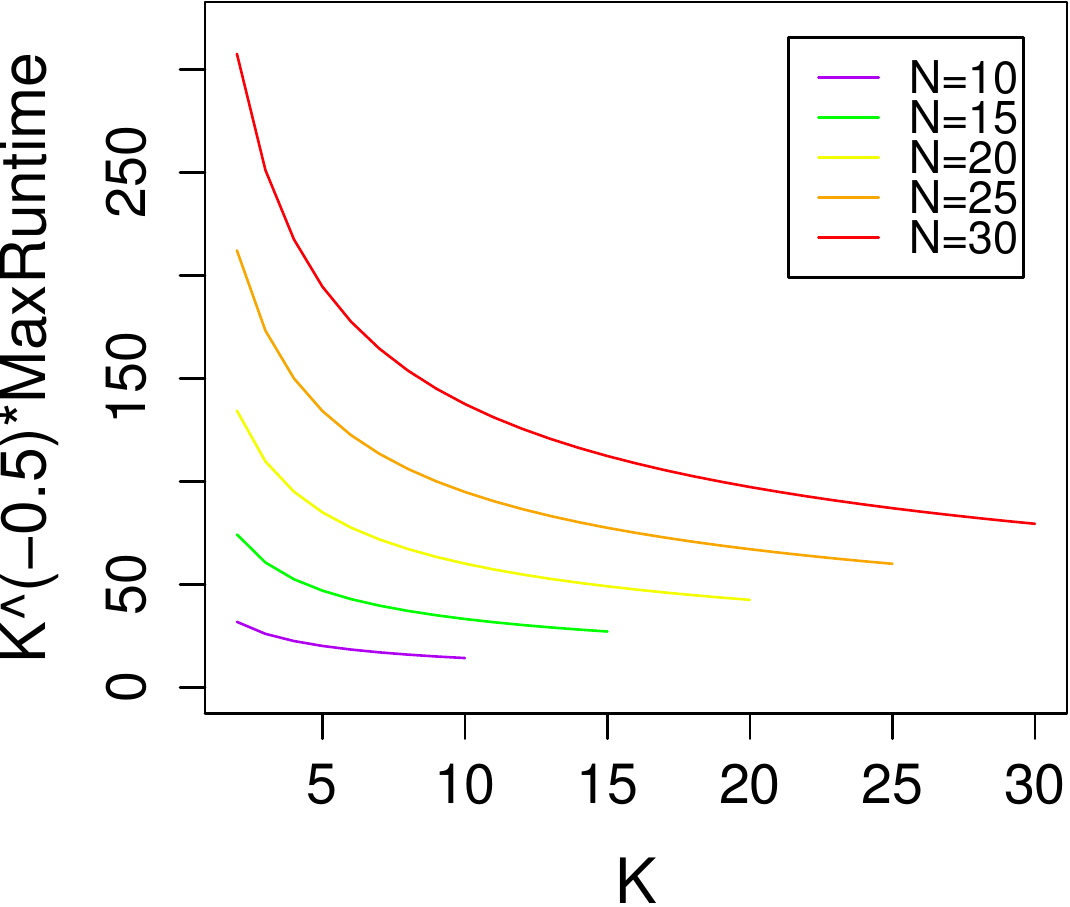}
}
\caption{The value of feature $\sqrt[2]{\frac{1}{K}}\times MaxRuntime$ in TLR-SC, when $2\leq K \leq N$, $N = 10,15,20,25,30$}
\label{NewFeature-K}
\end{figure}
In order to capture this relationship, we create a new feature that directly couples K and MaxRuntime (which is also influenced by N). We empirically found that $\sqrt[2]{\frac{1}{K}}\times MaxRuntime$ captures best this non-linear relationship. Using the new feature and linear regresison (the $lm$ R package), MAPE reduces from 20\% to 4.58\%. Figure \ref{NewFeature-K} shows the value of feature $\sqrt[2]{\frac{1}{K}}\times MaxRuntime$, given $2\leq K\leq N, N = 10,15,20,25,30 $.

Note that this feature is customized for Quicksort. As we showed earlier (Figure~\ref{Fix-N-Algorithms}), different algorithms have different shapes of SC, thus feature $\sqrt[2]{\frac{1}{K}}\times MaxRuntime$ is not suitable for other algorithms. Therefore, in order to build models for predicting the SC of other algorithms, we require a more general approach for capturing the relationship between SC and the features.

\subsubsection{Optimization of the Proposed Feature}

The new feature, expressed as $(K^{-0.5})\times MaxRuntime$, generally captures the shape of how SC decreases, while K increases. We can optimize this result by parameterizing it. Using two parameters \emph{a} and \emph{b}, with initial value $a=0, b=-0.5$, the new feature becomes 
\begin{equation}
((K+a)^{b})\times MaxRuntime
\end{equation}

We vary the value of parameters $a$ and $b$ to find the best combination. 
We have found that for \emph{lm} (with features $((K+2.2)^{-0.7})\times MaxRuntime$, N and K) trained on data with $10\leq N \leq 20, 2\leq K \leq N$ and tested on data of $N=40, 2\leq K \leq N$, the best combination is $a=2.2, b=-0.68$, with Mean Absolute Error (MAE) of 7.04.

Adding more parameters can improve the accuracy of our model, but may result in overfitting. Overfitting refers to fitting the training set very well, while failing to generalize to new test data. To check against this, we test the parameter combination on more test sets, where $N=40,60,80, 2\leq K \leq N$, while the training set remains $10\leq N \leq 20, 2\leq K \leq N$. The overall results show that the following parameter combination works well on all test sets $a=2.2, b=-0.7$ and the final feature used in \emph{lm} is 
\begin{equation}
((K+2.2)^{-0.7})\times MaxRuntime
\end{equation}

\subsubsection{Results}

\begin{figure}[h!]
\centerline{
\includegraphics[width=9cm]{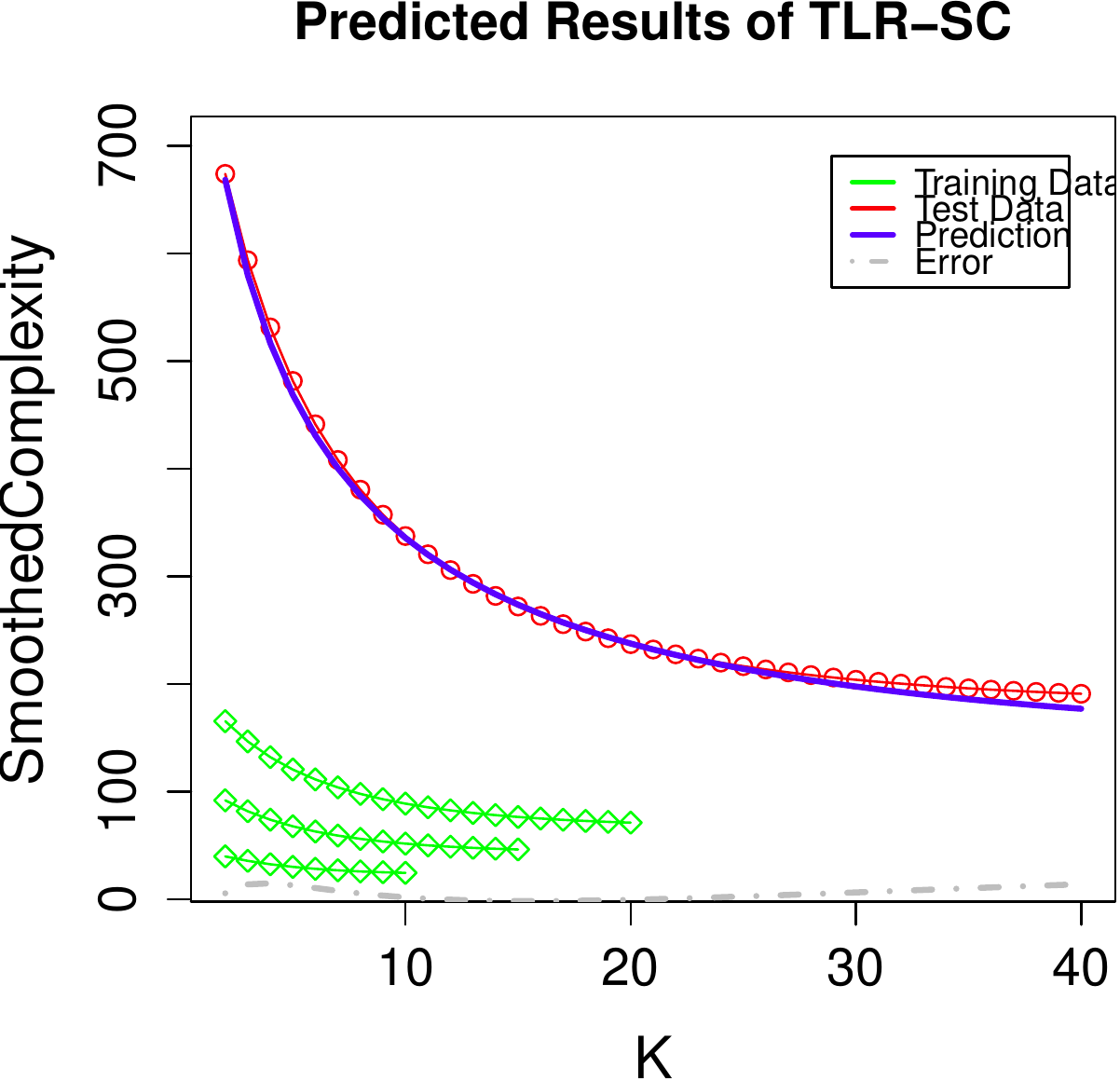}
}
\caption{The predicted results of TLR-SC, trained on data of $N=10,15,20, 2\leq K\leq N$, tested on data of $N=40, 2\leq K\leq N$.}
\label{Result-M1}
\end{figure}

We work with two ground truth datasets for training and testing models for Quicksort. The first set contains data of $10 \leq N \leq 100, 2 \leq K \leq N$ and N increases by 5. The second set contains data of $100 \leq N \leq 500, 2 \leq K \leq N$ and N increases by 100. Table \ref{table:Sample-M1} shows some sample data from the first dataset.
\begin{table}[h!]
\caption{Sample ground truth data for Quicksort used in TLR-SC.}
\label{table:Sample-M1}
\centerline{
\begin{tabular}{|c|c|c|c|} \hline\hline
N  & K & MaxRuntime & SC \\ \hline
10 & 2 & 45 & 39.7305 \\
10 & 3 & 45 & 35.6077 \\
10 & 4 & 45 & 32.4413 \\
10 & 10 & 45 & 24.4373 \\
15 & 2 & 105 & 91.8248 \\
15 & 3 & 105 & 81.6957 \\
15 & 4 & 105 & 73.8442 \\
...	&...	&... &...\\ \hline\hline
\end{tabular}
}
\end{table}
\begin{table}[h!]
\caption{Description of notation for TLR-SC.}
\label{table:Terms-M1}
\begin{center}
\begin{tabular}{|l|p{7.5cm}|} \hline\hline
Term  & Description \\ \hline\hline
$train_{a-b}$ & A training set of $a \leq N \leq b, 2\leq K \leq N$ \\\hline
$lm_{a-b}$ & An $lm$ model trained on $train_{a-b}$ \\\hline
$test_{a-b}$ & A test set of $a \leq N \leq b, 2\leq K \leq N$ \\\hline\hline
\end{tabular}
\end{center}
\end{table}
\begin{table}[htbp]
\caption{MAE and MAPE of the predicted results of TLR-SC.}
\label{table:Result-M1}
\centerline{
\begin{tabular}{|c|c|c|c|c|c|c|c|} \hline\hline
Model & Error &$test_{40-50}$& $test_{60-70}$ & $test_{90-100}$& $test_{200-200}$ &$test_{300-300}$& $test_{500-500}$ \\ \hline
\multirow{2}{*}{$lm_{10-20}$} & MAE & 7.85 & 17.58 & 38.83 & 183.02 & 422.28 &1175.83 \\\cline{2-8}
                              & MAPE & 2.56\% & 3.42\% & 4.41\% & 7.26\% & 9.61\% &13.66\% \\\hline\hline
\multirow{2}{*}{$lm_{10-40}$} & MAE & 4.26 & 12.49 & 34.22 & 200.44 & 483.83 &1377.71 \\\cline{2-8}
                              & MAPE & 1.33\% & 2.44\% & 3.88\% & 7.85\% & 10.91\% &15.99\% \\\hline\hline
\multirow{2}{*}{$lm_{10-60}$} & MAE & 2.73 & 8.53 & 28.80 & 200.08 & 497.88 &1442.20 \\\cline{2-8}
                              & MAPE & 0.67\% & 1.56\% & 3.15\% & 7.54\% & 10.87\% &16.38\% \\\hline\hline
\multirow{2}{*}{$lm_{10-80}$} & MAE & 5.26 & 4.98 & 20.89 & 185.03 & 480.79 &1432.44 \\\cline{2-8}
                              & MAPE & 1.47\% & 0.71\% & 2.23\% & 6.78\% & 10.24\% &15.98\%  \\\hline\hline
\end{tabular}
}
\end{table}
We denote by $train_{a-b}$ a training set with $a \leq N \leq b, 2\leq K \leq N$, $test_{a-b}$ a test set with $a \leq N \leq b, 2\leq K \leq N$, and $lm_{a-b}$, an $lm$ model trained on $train_{a-b}$. These notations are listed in Table \ref{table:Terms-M1}.
Figure \ref{Result-M1} shows the predicted results of $lm_{10-20}$ on $test_{40-40}$, the values of data in $train_{10-20}$, and the true value of the SC of Quicksort in $test_{40-40}$.

Table \ref{table:Result-M1} shows the MAE and MAPE of $lm$ trained on different training sets and tested on various test sets. Generally speaking, the larger the training set, the better the test results, and the greater the N values in test set, the worse the prediction accuracy. For test set with small N values (e.g., $N\leq 100$), MAPE is around 3\%. However, when tested on greater N values (e.g., $N \geq 200$), $MAPE \geq 10\%$.

\subsubsection{Discussion}

The main idea behind TLR-SC is to transfer a nonlinear regression problem into a linear one, by manipulating data and features. The advantage is that we can use a simple algorithm - linear regression, to handle a complex data shape in one step, and therefore the TLR-SC model is really simple. This model shows the importance of analyzing the data and understanding the relationship between features and prediction target. No matter how powerful the learning algorithm (e.g., WEKA algorithms), it cannot automatically solve the problem without human knowledge.

Although the TLR-SC results are very encouraging, the accuracy drops quickly when the N value of the test set is large. One reason might be that the training is not enough, so the model cannot show its full ability in learning and predicting. It is difficult to expect it performing well on test sets with $N \geq 300 $, when we only train it on data of $N \leq 20 $. Besides, transforming the data and features to create a linear relationship may cause the prediction error of linear regression to be distorted \citep{motulsky2004fitting}. 
Another reason is that possibly other functions may work better than our choice $K^{b}$. Most importantly, as discussed in Section~\ref{sec:dataanalysis}, the relationship between SC and K is changing, given different values of N. The greater N is, the quicker SC decreases when K increases. Unfortunately, our customized feature does not capture this well. Although the value of feature $((K+a)^{b})\times MaxRuntime$ changes with N, the parameters \emph{a} and \emph{b} do not. However, parameter \emph{b} is the most important one in fitting the curve, and if we can make \emph{b} change based on N, the performance of TLR-SC should improve. In the next section, we present a new model that implements this observation.

Another disadvantage of TLR-SC is that it is hard to transfer to other algorithms. The feature $((K+2.2)^{-0.7})\times MaxRuntime$ is created to fit the data shape of Quicksort. Other algorithms may have a totally different shape, therefore the model may not be applicable to other algorithms by solely changing the value of parameters. The biggest problem with the transition is that for other algorithms, we lack the necessary amount of data to analyze the data shape, so it is difficult to find the right function shape.

\subsection{Model NLR-SC}

In this section, we propose a new model for predicting the SC, {\bf NLR-SC} (Non-linear Regression for Smoothed Complexity). This is an updated model, aimed at solving the problems of TLR-SC. We previously showed that by carefully capturing the nonlinear data shape, TLR-SC can dramatically increase the accuracy of prediction. Nevertheless, the relationship between SC and the perturbation parameter K is changing, for different values of N, and the fixed parameters \emph{a} and \emph{b} limit the ability of TLR-SC to capture this changing shape, as well as to be transferred to other sorting algorithms.

To solve these problems, NLR-SC breaks down the surface-fitting problem into multiple curve-fitting problems. By predicting the SC curve by curve, NLR-SC gradually predicts the whole surface. 
\begin{figure}[htbp]
\begin{center}
\includegraphics[trim = 0.5cm 1.0cm 1cm 2.5cm, clip,width=10cm]{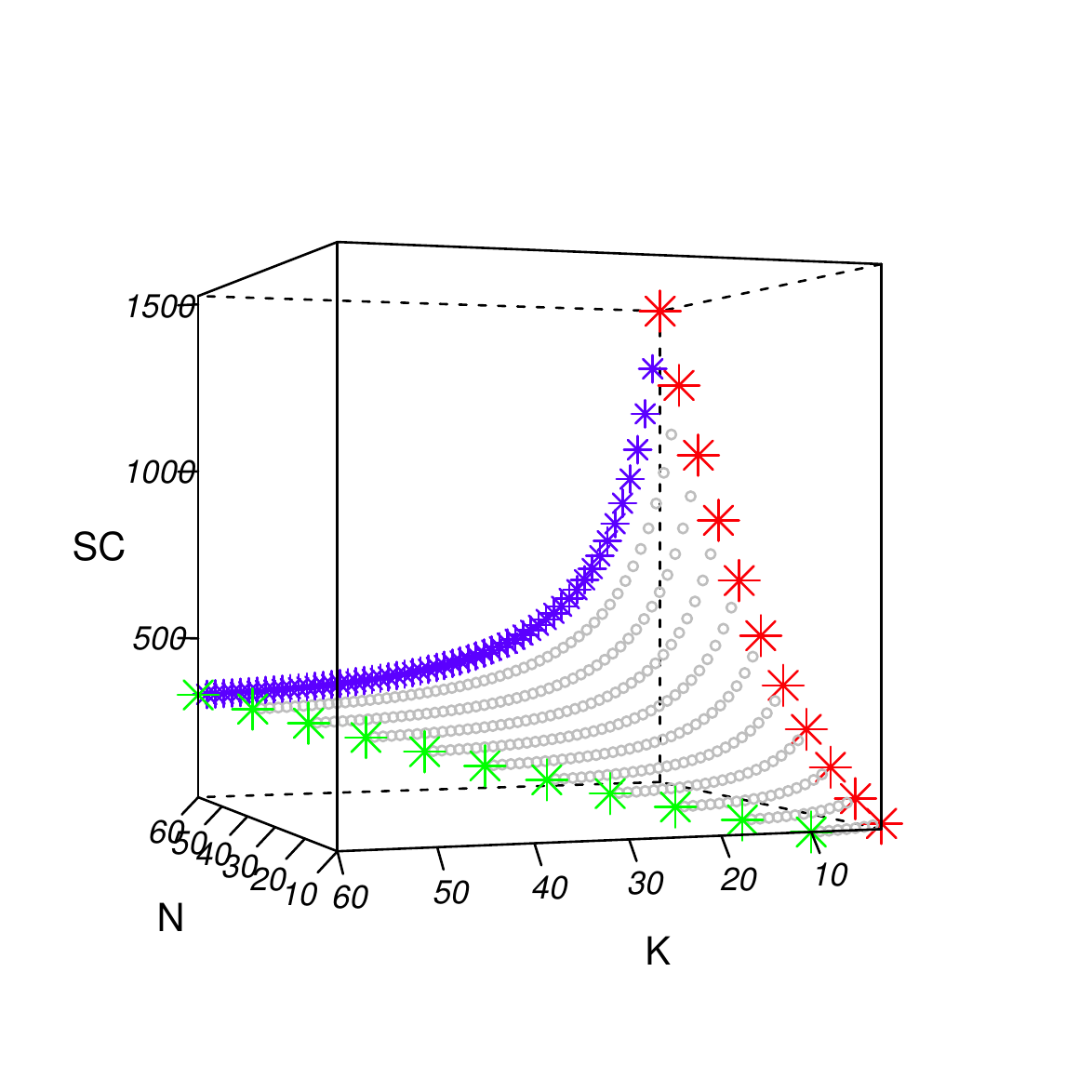}
\caption{Three curves fitted in NLR-SC. Blue curve is the SC of fixed N and varying K; Red curve is the SC of varying N and fixed $K =2$; Green curve is the SC of varying N and fixed $K=N$.}
\label{ThreeCurves}
\end{center}
\end{figure}
We deal with three type of curves in NLR-SC, first one, the curve of SC and K, for fixed N, is shown in blue in Figure \ref{ThreeCurves}. Because the shape between the SC and K changes for different N, we divide the surface into curves by N values, and predict these curves one by one. For instance, to predict the SC of $40\leq N\leq 50, 2\leq K \leq N$, we first predict the SC of $N=40, 2\leq K \leq N$, then $N=41, 2\leq K \leq N$, then $N=42, 2\leq K \leq N$, \dots, until $N=50, 2\leq K \leq N$. For each N, we re-calculate the parameters in the fitting function, so that we capture the data shape of the SC. For this, we employ $nls$, an R model fitting library, which automatically determines the nonlinear least-squares estimates of the parameters. $nls$ works well when the function shape is decided, but the parameters of the function are uncertain. We refer to $nls$ predicting the first type of curve as sub model NLR-SC-N.

By fixing K to 2 or N, we can get two more curves of the SC and N, shown in red and green in Figure \ref{ThreeCurves}. When we try to use data of small N and K, to predict the SC of large N and K, these two curves are bridges between training data and the predicted target, and they define the starting point and the ending point of curve one. The reason for fixing K to 2 and N is that, according to the theory of smoothed analysis, we know that when $K=2$, the curve of the SC and N follows the worst-case behaviour, while when $K=N$, the curve follows the average-case behaviour. Therefore, we can use two $lm$ models to capture these two curves well, and supply NLR-SC-N with their prediction as a training set. We refer to this part as submodel NLR-SC-K.

\begin{figure}[htbp]
\centerline{
\includegraphics[trim = 0cm 0cm 0cm 0cm, clip,width=14cm]{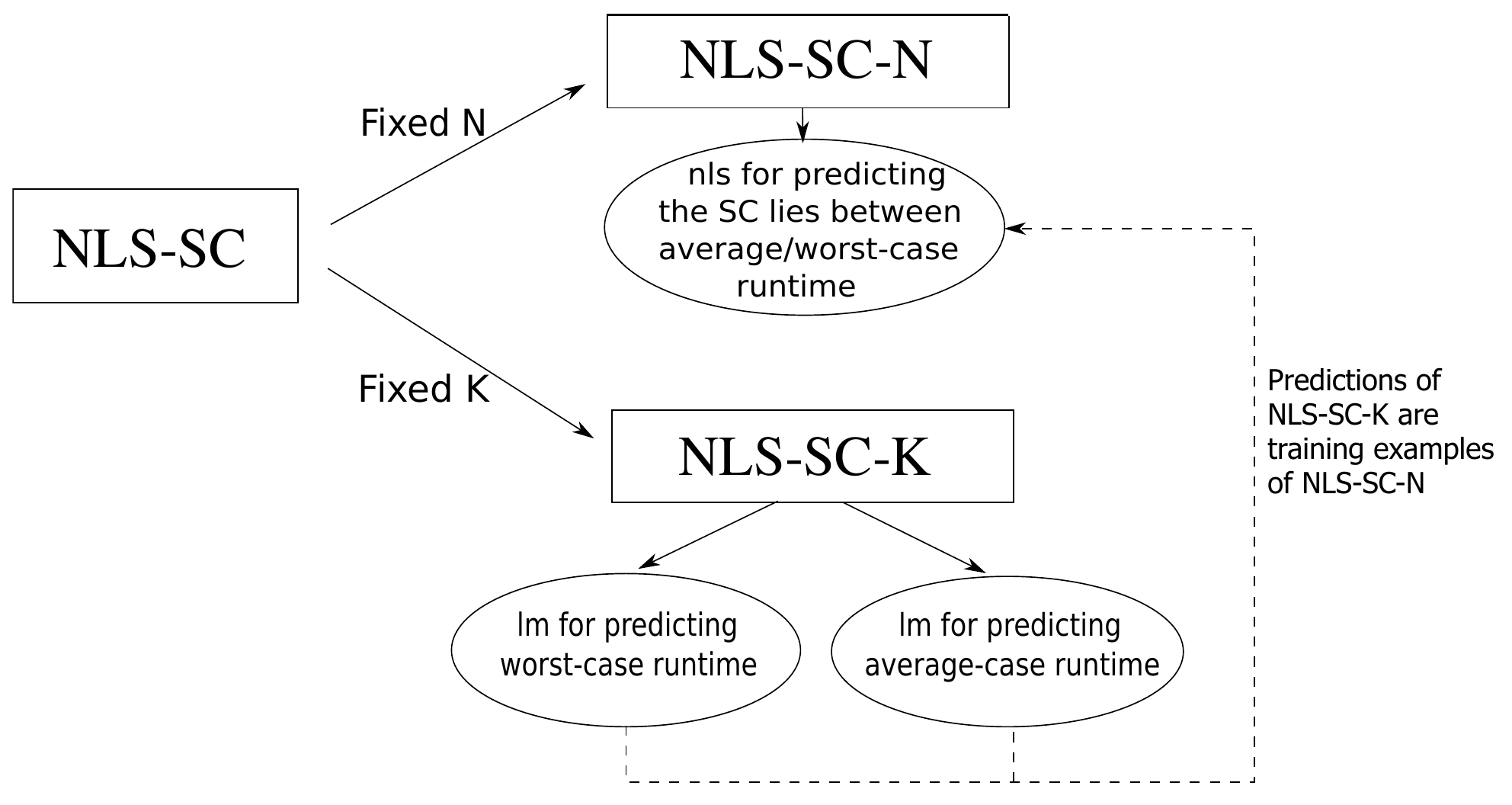}
}
\caption{The structure of model NLR-SC.}
\label{Model2}
\end{figure}

NLR-SC is the combination of NLR-SC-N and NLR-SC-K, shown in Figure \ref{Model2}. 
By fixing either N or K, we simplify the data shape and make the hidden patterns explicit. 

\subsubsection{NLR-SC: Fixed N}

\begin{figure}[h!]
\centerline{
\includegraphics[trim = 0cm 0.5cm 0cm 0.5cm, clip,width=14cm]{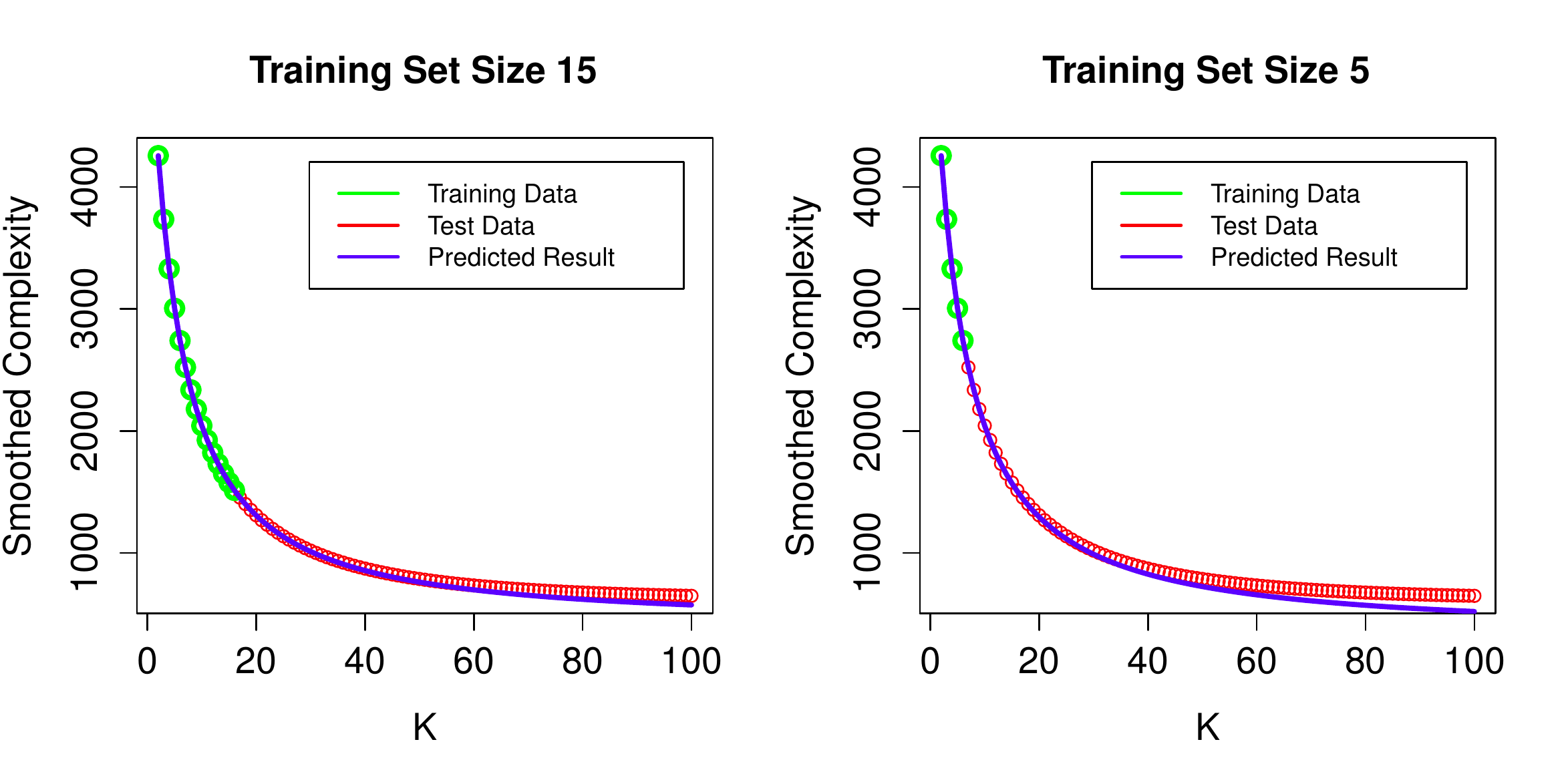}
}
\caption{Testing the size of the training set that $nls$ needs to deliver good results, for $N=100,2\leq K\leq 16$ and for $N=100,2\leq K\leq 6$}
\label{TrainSet-M2}
\end{figure}

We predict the SC of data of fixed N, using $nls$. 
The function we fit with $nls$ is 
\begin{equation}\label{function-M2}
a(\frac{K}{N}+c)^b+d
\end{equation}
\noindent where \emph{a}, \emph{b}, \emph{c}, \emph{d} are parameters to be fitted. Equation \ref{function-M2} is inspired by the feature used in TLR-SC: 
\begin{equation}\label{function-M1}
  ((K+2.2)^{-0.7})\times MaxRuntime
\end{equation}
Compared to Equation \ref{function-M1}, in Equation \ref{function-M2}, we replace the \emph{MaxRuntime} by \emph{a} to increase the flexibility. In addition, we chose $\frac{K}{N}$ instead of $K$ in Equation \ref{function-M2}, because the maximum value of $K$ depends on the value of $N$, and we need the $nls$ model to work on all $N$ values, therefore it is better to use the proportion of K to N, rather than the absolute value of K.
We ran an experiment to examine the minimum training data that $nls$ needs, to deliver good results. Figure \ref{TrainSet-M2} shows how $nls$ performs when trained on data of $N=100, 2\leq K\leq 16$  and $N=100, 2\leq K\leq 6$ separately. Green spots in Figure \ref{TrainSet-M2} are training data, red spots are test data, and the blue line is the predicted result. 
\begin{figure}[h!]
\begin{center}
\includegraphics[width=9cm]{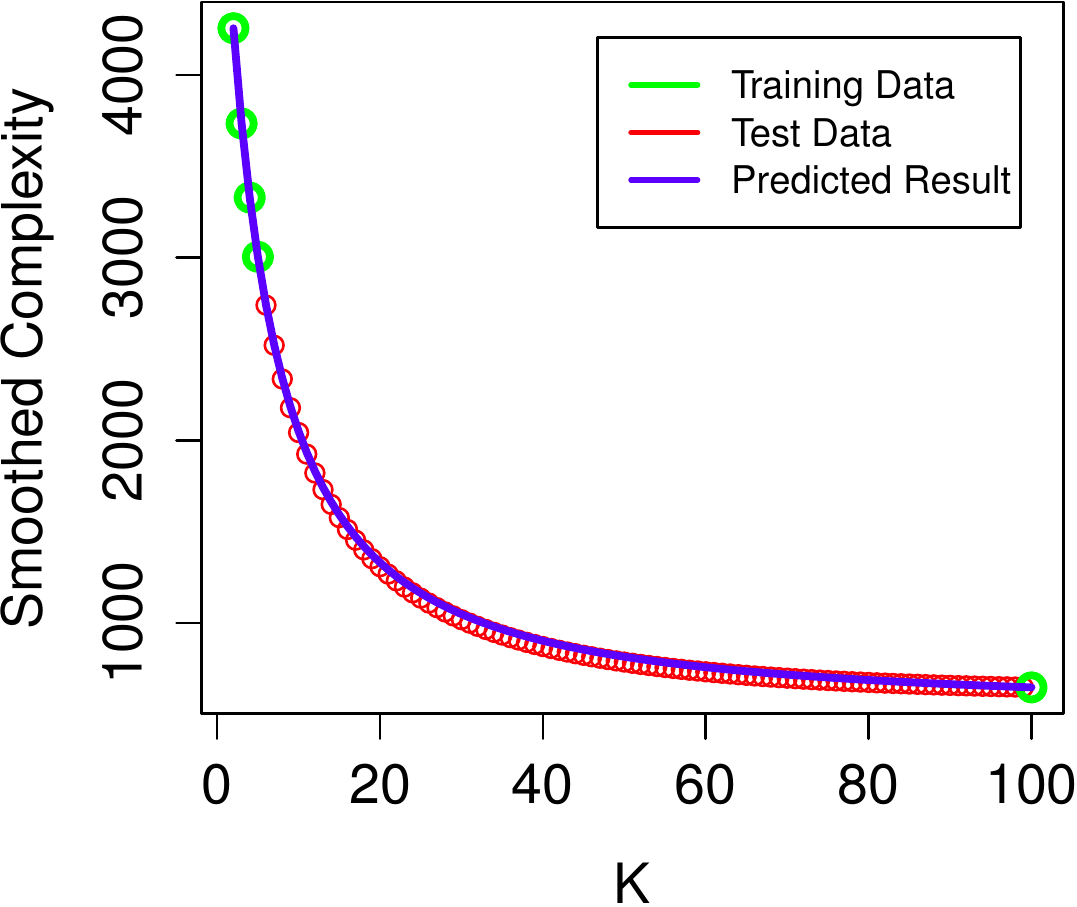}
\caption{By adding one important training example $N=100, K=N$, $nls$ can find the perfect fit to the curve with only 5 training cases.}
\label{TrainSet-5-M2}
\end{center}
\end{figure}

We see from Figure \ref{TrainSet-M2} that the performance of $nls$ is already very good with only 5 training examples. This is likely due to the fact that Equation \ref{function-M2} fits the data well. The main reason why the accuracy decreases when we reduce the the training set is that the algorithm does not get enough information about where the SC will decrease to, in other words, what is the value of the SC when $N=100, K=N$. By adding the data entry of $N=100, K=N$ into the training set, the prediction accuracy is greatly improved. Shown in Figure \ref{TrainSet-5-M2}, with one extra data entry in the training set, $nls$ easily finds the perfect fit to the curve.

\begin{figure}[htbp]
\centerline{
\includegraphics[width=14cm,height = 15cm]{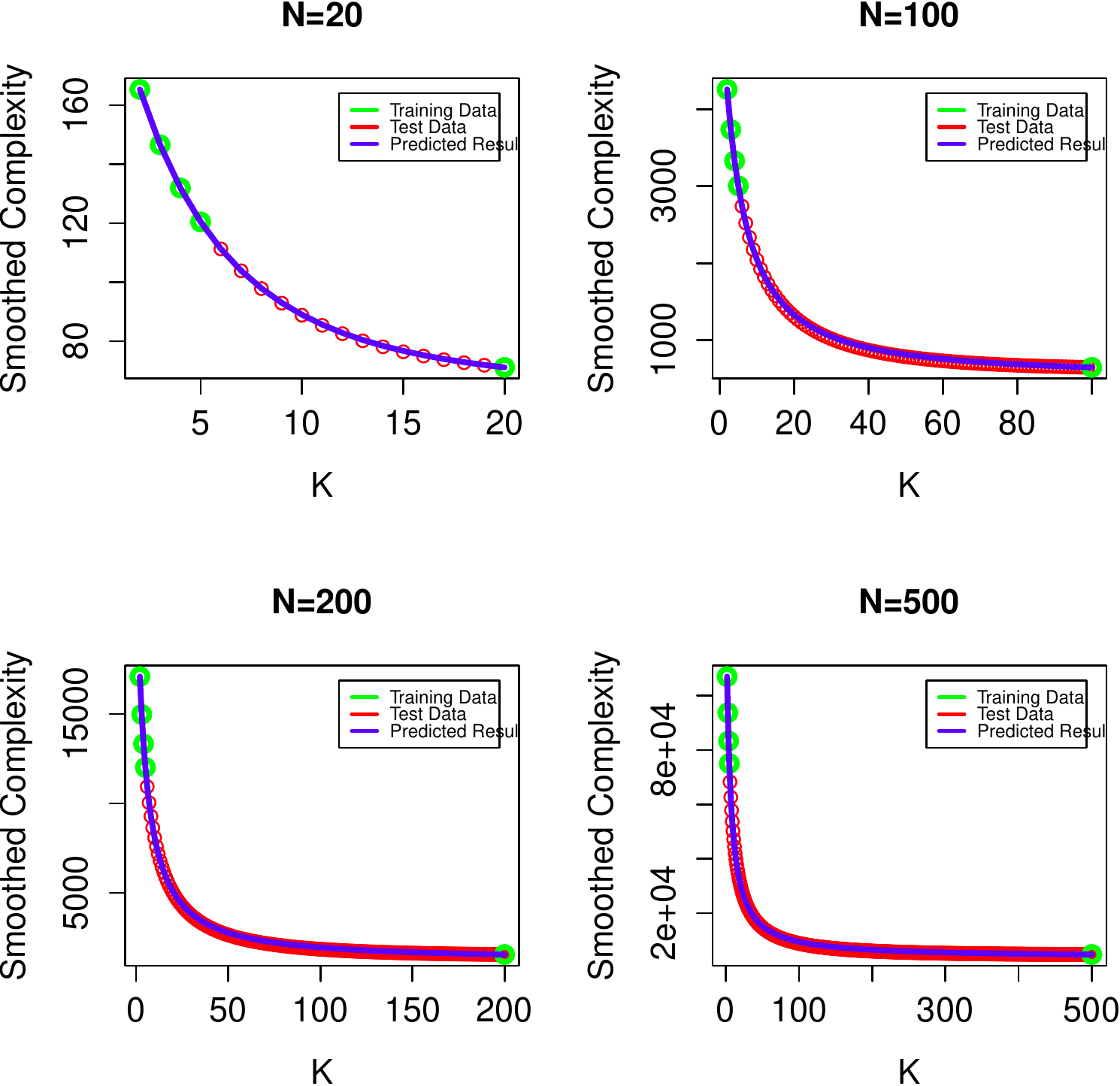}
}
\caption{Predicting the SC with 5 training examples, for $N=20,100,200,500$, $2\leq K\leq N$}
\label{TrainSet-Result-M2}
\end{figure}

As shown in Figure \ref{TrainSet-Result-M2}, examining data with $N = 20,100,200,500$, as long as the training data contains entries whose $K=2$, and $K=N$, $nls$ can find the perfect fit using only 5 training cases. Note that the size of the training set must be larger than the number of parameters in $nls$, and more parameters generally means higher prediction accuracy.

Although both submodel NLR-SC-N of NLR-SC, and model TLR-SC, deal with the same nonlinear relationship, they take two different approaches. TLR-SC solves a surface fitting problem, using training data of small N and K, predicting the SC for large N and K, while NLR-SC-N solves a curve fitting problem, using training data of a specific N and some K values, predicting the SC of same N and other K values. The NLR-SC-N can be used to solve the same surface fitting problem as TLR-SC does, by dividing the surface into multiple curves by N, and solving them one by one. Because, for each N value, a new curve is constructed by $nls$, NLR-SC-N can capture the changing shape between the SC and K more accurately than TLR-SC, without causing overfitting. 

The only problem now is how to generate training data for NLR-SC-N. To predict the SC of a specific N value, $nls$ needs at least 5 training examples of same N value, $K=2$, $K=N$ and other K values. If N is large, we cannot collect such training data through an experimental approach. Therefore, we use submodel NLR-SC-K to predict the required training data for NLR-SC-N.

\subsubsection{NLR-SC: Fixed K}

\begin{figure}[htbp]
\begin{center}
\includegraphics[trim = 0cm 0.5cm 0cm 1cm, clip,width=9cm]{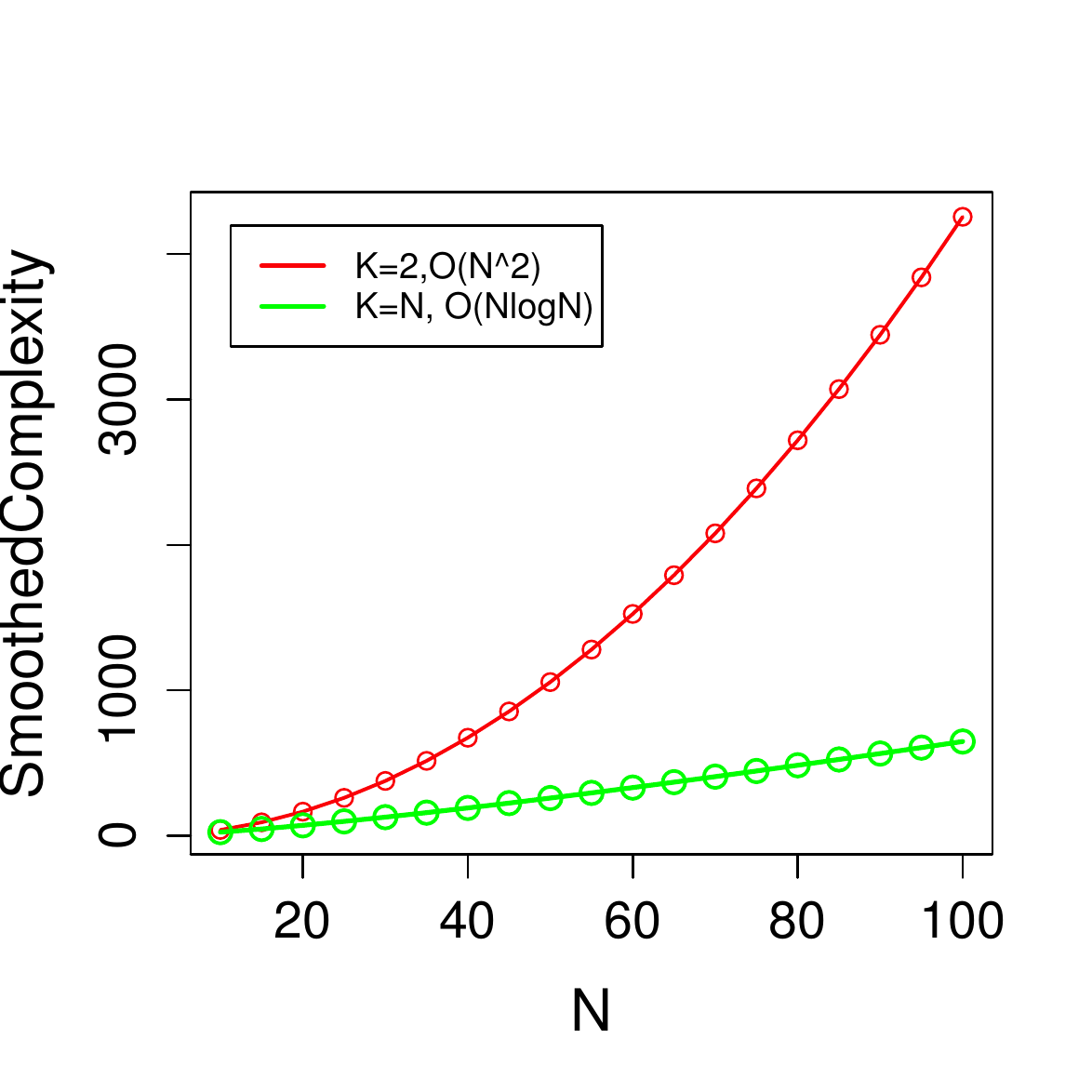}
\caption{The SC of Quicksort when $K=2$ and $K=N$, $5\leq N\leq 100$.}
\label{FixK-SC-N-M2}
\end{center}
\end{figure}

Submodel NLR-SC-K is created to generate the training data for $nls$ in NLR-SC-N. As explained in the previous section, $nls$ needs two important training data points, whith $K=2$ and $K=N$, to determine the starting point and the ending point of the curve. As shown in Figure \ref{FixK-SC-N-M2}, when $K=2$, the SC of Quicksort increases while N increases, following the worst-case behavior $O(N^2)$. Similarly, when $K=N$, the SC follows the average-case behavior $O(N\log N)$. Both experimental and theoretical results support this finding.

\begin{figure}[htbp]
\begin{center}
\includegraphics[trim = 0cm 0.5cm 0cm 1cm, clip,width=9cm]{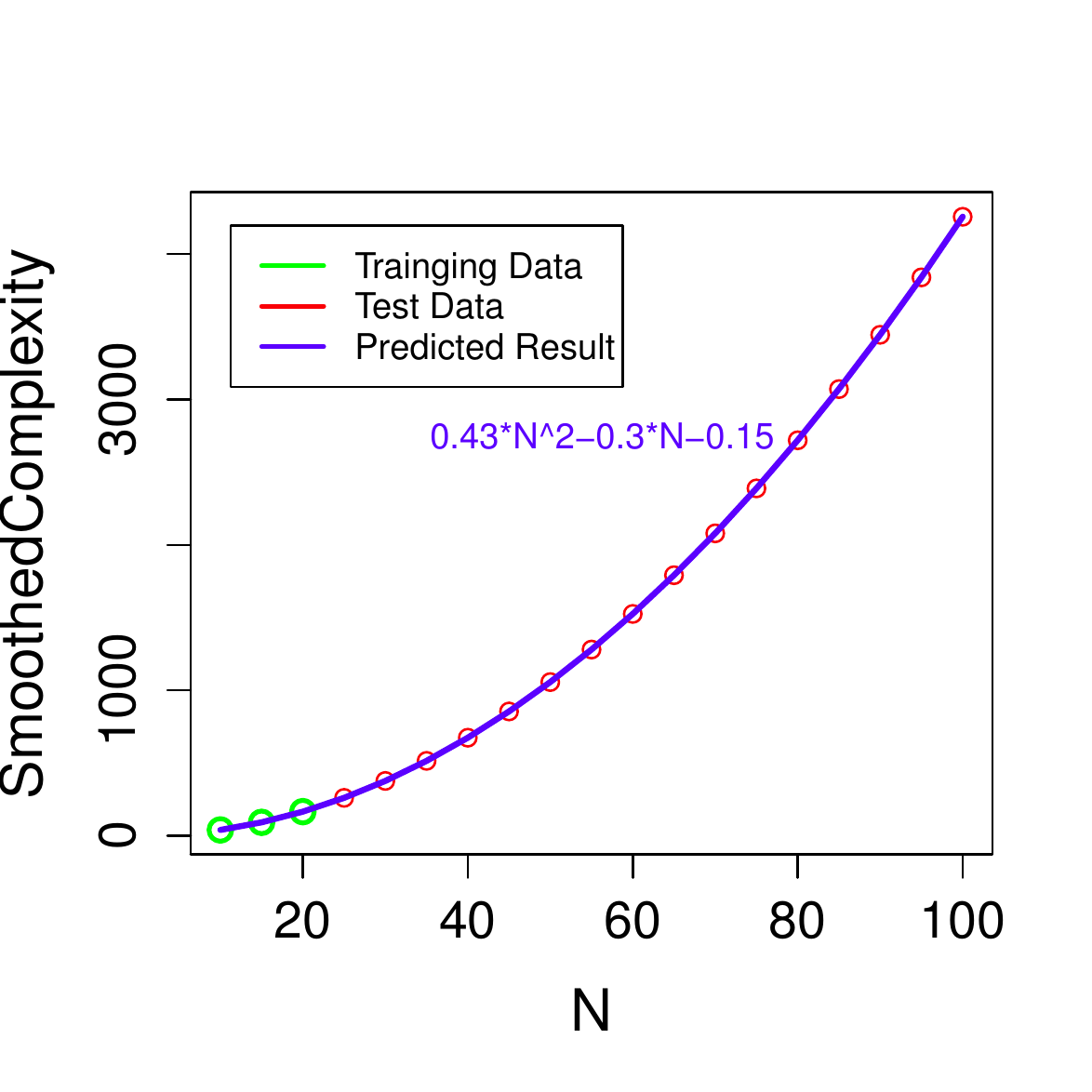}
\caption{Predicting the worst-case complexity of Quicksort, by using $lm$ that was trained on data of $N\leq 20$, $K=N$ and tested on data of $N > 20$, $K=N$}
\label{FixK-2-Result-M2}
\end{center}
\end{figure}

\begin{figure}[htbp]
\begin{center}
\includegraphics[trim = 0cm 0.5cm 0cm 1cm, clip,width=9cm]{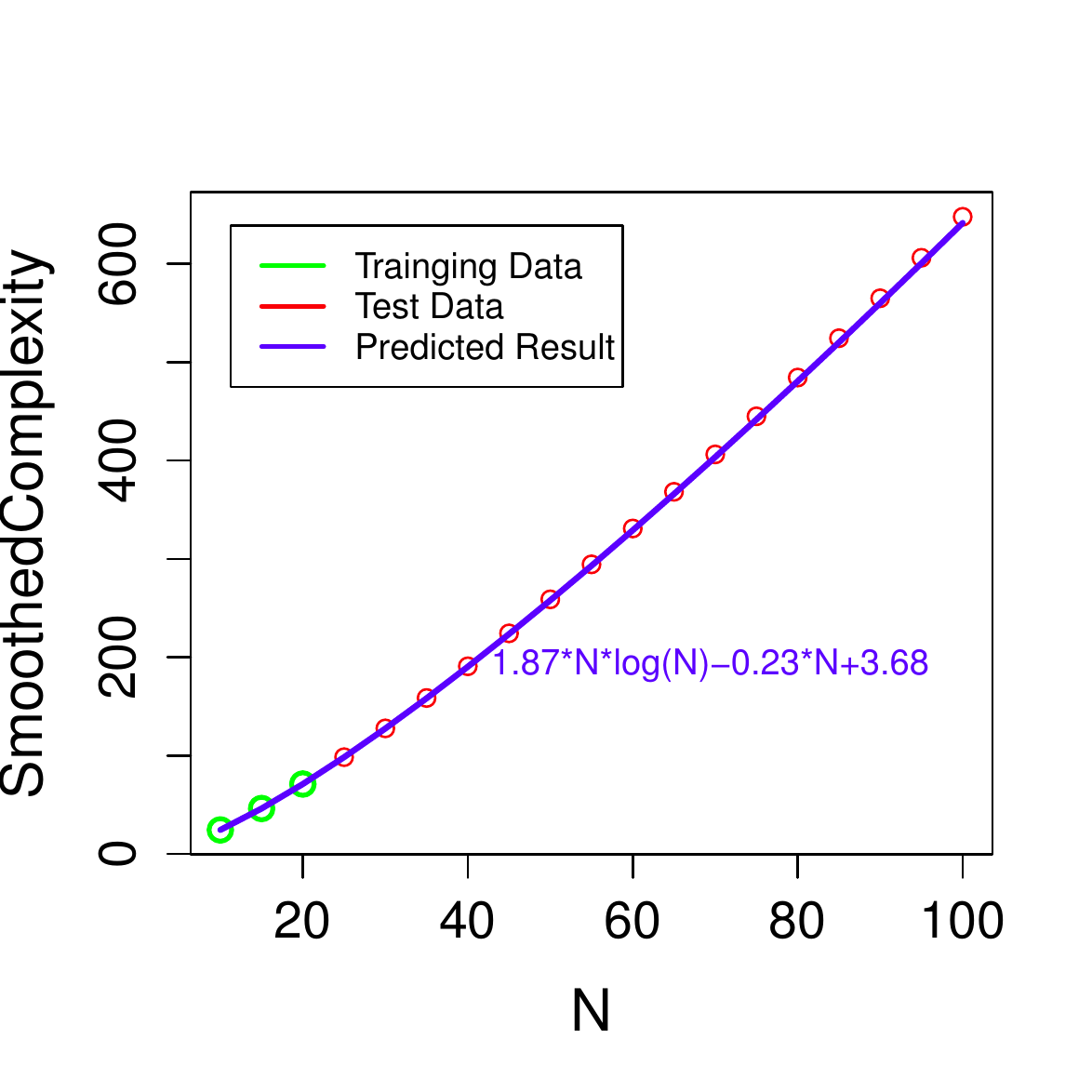}
\caption{Predicting the average-case complexity of Quicksort, by using $lm$ that was trained on data of $N\leq 20$, $K=N$ and tested on data of $N > 20$, $K=N$}
\label{FixK-N-Result-M2}
\end{center}
\end{figure}

These two patterns are so clear that we can use two $lm$ models to fit the curves. One model focuses on $K=2$: we use $N^2$ and $N$ as features in $lm$, train it on data with small $N$, and test it on data with big $N$. Similarly, we fit another $lm$ model for $K=N$: we use $N\times  log(N)$ and $N$ as features. Results are shown in  Figure \ref{FixK-2-Result-M2} and Figure \ref{FixK-N-Result-M2}.
Besides $K=2$ and $K=N$, the SC of K values close to 2 or N can also be predicted. For example, the $lm$ model which is applicable for $K=2$, is also applicable for $K=3, K=4,$ and $ K=5$. Similarly, the $lm$ model which is applicable for $K=N$, is also applicable for $K=N-1, K=N-2$, and $ K=N-3$. However, the accuracy of such predicted results might be compromised a little, and it is better to keep the value of K close to 2 and N.
By using these two $lm$ models, we can generate the necessary training cases for NLR-SC-N.

\subsection{Combining the NLR-SC-N and NLR-SC-K}

NLR-SC is the combination of NLR-SC-N and NLR-SC-K. NLR-SC-N is used to predict the SC of fixed large N, using $nls$, and NLR-SC-K is used to produce training data for NLR-SC-N, using $lm$.

\begin{figure}[htbp]
\begin{center}
\includegraphics[width=9cm]{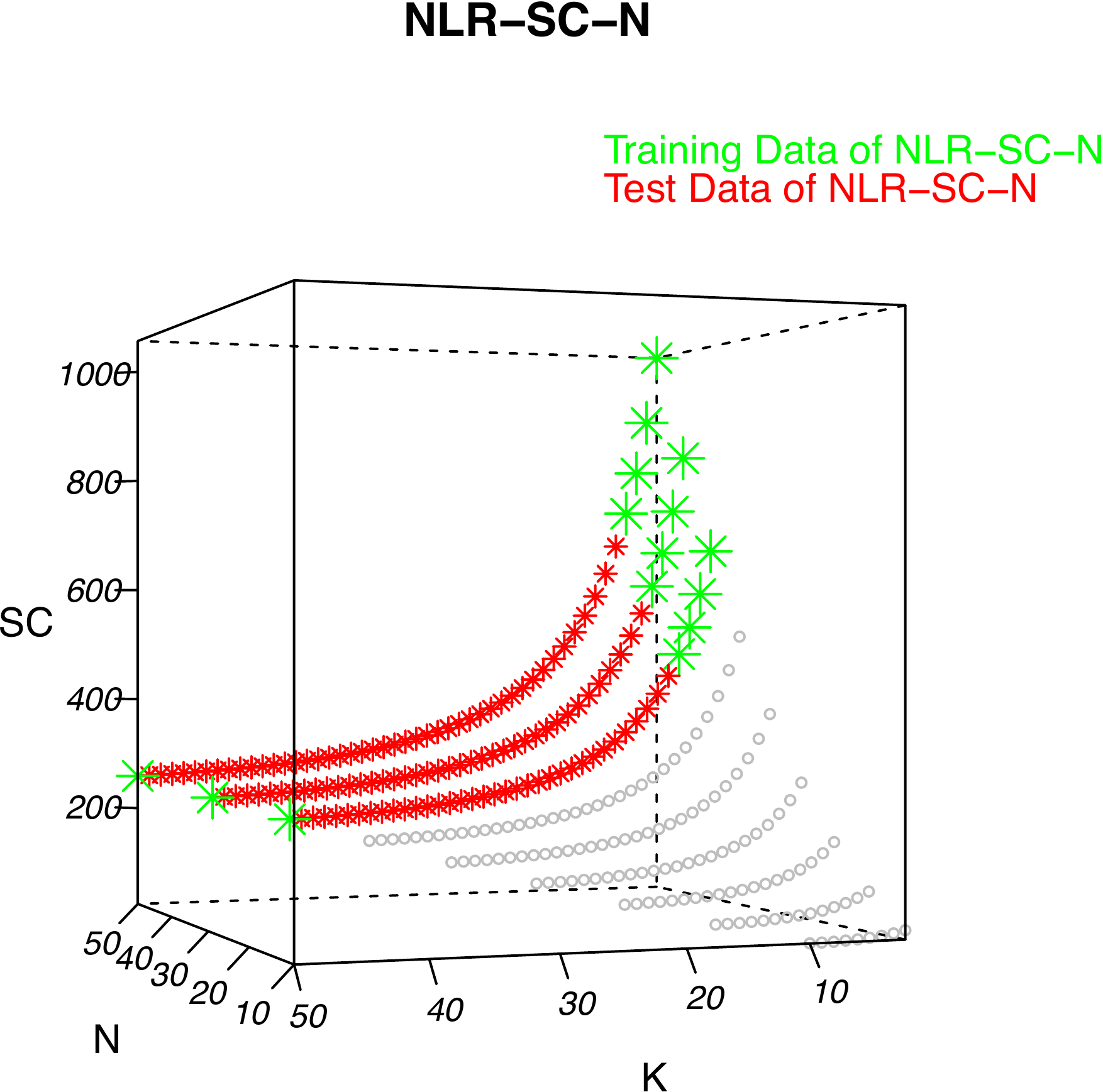}
\caption{The range of training and test data of NLR-SC-N}
\label{3D-Submodel-I}
\end{center}
\begin{center}
\includegraphics[width=9cm]{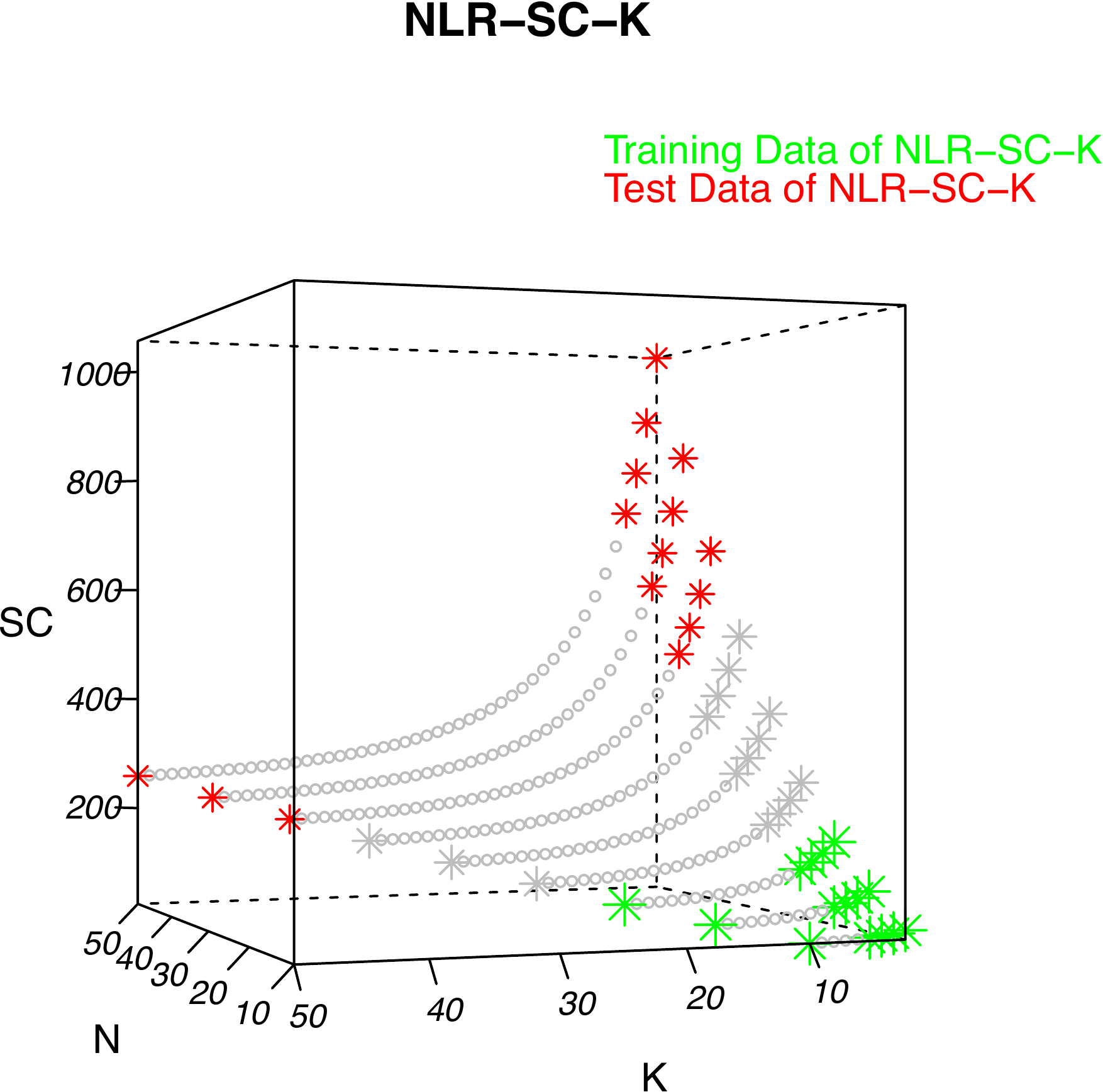}
\caption{The range of training and test data of NLR-SC-K}
\label{3D-Submodel-II}
\end{center}
\end{figure}

In NLR-SC-N, for each N value we want to predict the SC for, we create a standalone $nls$ model. Figure \ref{3D-Submodel-I} shows 3 $nls$ models for predicting the SC of $N=40,45,50$, $6\leq K \leq N-1$. Each $nls$ is trained on 5 training examples, marked as green stars, and these 3 $nls$ models work one by one to predict the target surface of $40\leq N \leq 50$, $6\leq K \leq N-1$. 
In NLR-SC-K, for each K value we want to predict the SC for, we also create a standalone $lm$ model. Figure \ref{3D-Submodel-II} shows 5 $lm$ models for predicting the SC of $40\leq N \leq 50$, $K=2,3,4,5,N$. Each $lm$ is trained on 3 training examples of $N=10,15,20$ and same K value as the prediction target, marked as green stars. These 5 $lm$ models also work one by one to predict the SC of $40\leq N\leq 50$, $K=2,3,4,5,N$.
\begin{figure}
\centerline{
\includegraphics[width=9cm]{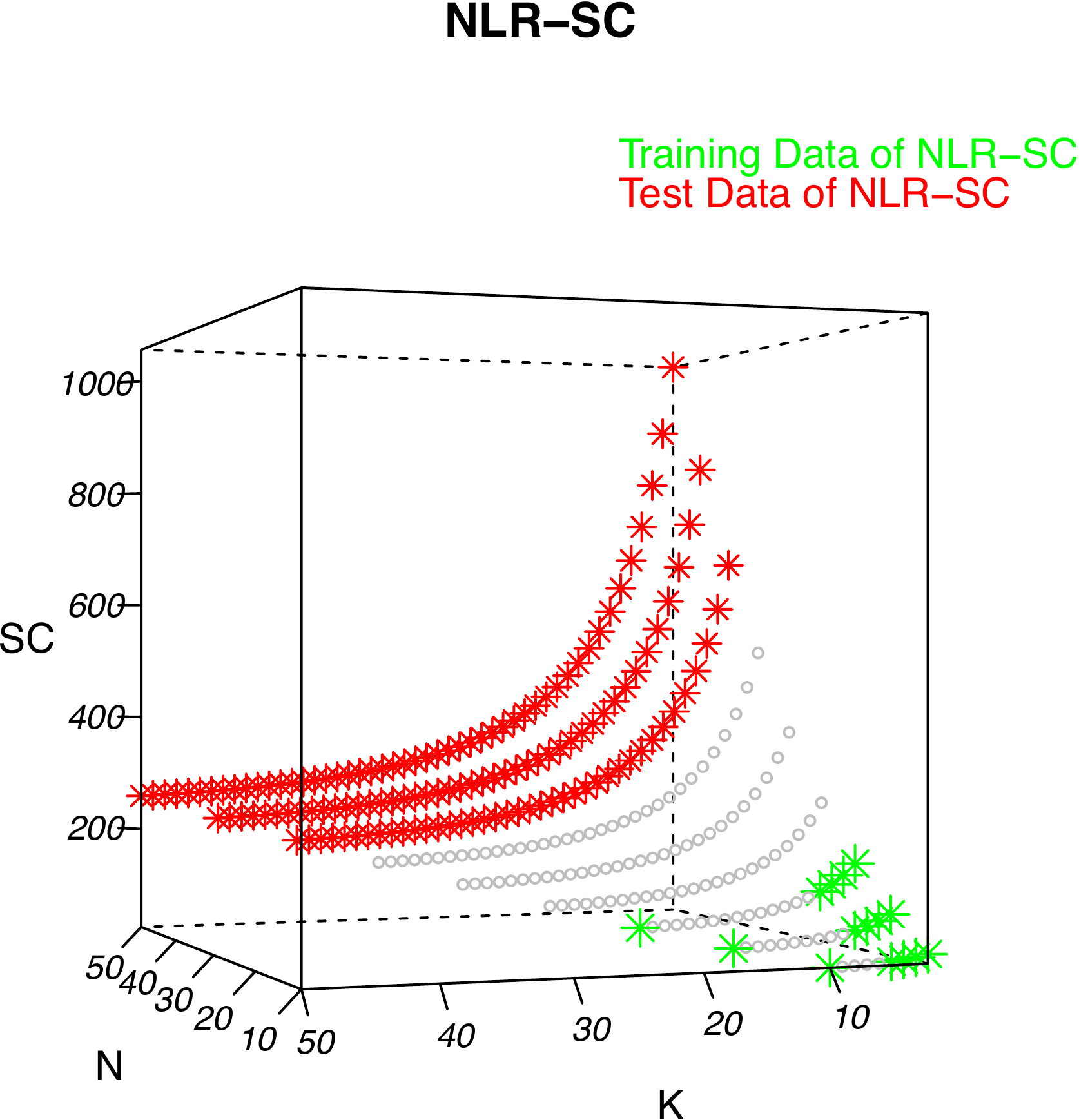}
}
\caption{The range of training and test data of NLR-SC, when combining NLR-SC-N and NLR-SC-K.}
\label{3D-Model-II}
\end{figure}
The most important point of the combination is that the training examples of NLR-SC-N are the predicted results of NLR-SC-K. In Figure \ref{3D-Submodel-II}, the red stars representing the predicted result of NLR-SC-K, change to green in Figure \ref{3D-Submodel-II}, indicating they are the training examples of NLR-SC-N. In other words, NLR-SC-N is built on the predicted results of NLR-SC-K. 

Figure \ref{3D-Model-II} shows the final combination: NLR-SC. When we consider NLR-SC-N and NLR-SC-K together, the training set of the whole model is data of $10\leq N\leq 20$, $K=2,3,4,5,N$, and the prediction target is data of $40\leq N\leq 50$, $2\leq K\leq N$, marked as green and red stars in Figure \ref{3D-Model-II}. This demonstrates the ability of NLR-SC to use training data of small N and K to predict the data of big N and K (as we will show, NLR-SC also works well for $N=3000$). 
We give the detailed predicting process to explain the combination. As shown in Figure \ref{3D-Model-II}, suppose the training data is the SC of Quicksort for $N=10,15,20$, $2 \leq K \leq N $, and the test data is the SC of Quicksort for $N= 40,45,50$, $2 \leq K \leq N $. 
First, we filter the training data into a smaller dataset $Train_{K=2}$, by $K=2$, $N=10,15,20$ and also filter the test data to a smaller dataset $Test_{K=2}$, by $K=2$, $N=40,45,50$. Table \ref{table:Sample-K-2} shows the data in $Train_{K=2}$ and $Test_{K=2}$. Next, we train $lm$ in NLR-SC-K on $Train_{K=2}$, using features $N^2$ and $N$, and test it on $Test_{K=2}$. We calculate the prediction error, and prepare the predicted results into different training sets for NLR-SC-N: $Train_{N=40}$, $Train_{N=45}$, and $Train_{N=50}$, by N value. We repeat this process for $K=3,4, 5$ and $K=N$. Note that for $K=N$, we change the features in $lm$ to $N\times log(N)$ and $N$.
\begin{table}[htbp]
\caption{The training/test sets and the predicted results of NLR-SC-K, when $K=2$.}
\label{table:Sample-K-2}
\centerline{
\begin{tabular}{|c|c|c|} \hline
\multicolumn{3}{|c|}{$Train_{K=2}$}\\\hline\hline
N & K & SC \\\hline
10 & 2 & 39.7305 \\
15 & 2 & 91.8248 \\
20 & 2 & 165.3564 \\
\hline\hline
\end{tabular}
\quad
\begin{tabular}{|c|c|c|} \hline
\multicolumn{3}{|c|}{$Test_{K=2}$}\\\hline\hline
N & K & SC \\\hline
40 & 2 & 673.8097 \\
45 & 2 & 854.5003 \\
50 & 2 & 1056.6209 \\
\hline\hline
\end{tabular}
\quad
\begin{tabular}{|c|c|c|} \hline
\multicolumn{3}{|c|}{Predicted Result}\\\hline\hline
True SC & Pred SC & AbsError \\\hline
673.8097 & 673.8558 & 0.0461 \\
854.5003 & 854.5739 & 0.0736 \\
1056.6209 & 1056.7293 &  0.1084\\
\hline\hline
\end{tabular}
}
\end{table}

\begin{table}[htbp]
\caption{The training/test sets and the predicted results of NLR-SC-N, when $N=40$.}
\label{table:Sample-N-40}
\centerline{
\begin{tabular}{|c|c|c|} \hline
\multicolumn{3}{|c|}{$Train_{N=40}$}\\\hline\hline
N & K & SC \\\hline
40 & 2 & 673.8558 \\
40 & 3 & 593.7202  \\
40 & 4 & 531.3032  \\
40 & 5 & 481.5961  \\
40 & 40 & 190.3537  \\
\hline\hline
\end{tabular}
\quad
\begin{tabular}{|c|c|c|} \hline
\multicolumn{3}{|c|}{$Test_{N=40}$}\\\hline\hline
N & K & SC \\\hline
40 & 6 & 441.1757 \\
40 & 7 & 408.0416 \\
40 & 8 & 380.4325 \\
... & ... & ... \\
40 & 39 & 191.6724 \\
\hline\hline
\end{tabular}
\quad
\begin{tabular}{|c|c|c|} \hline
\multicolumn{3}{|c|}{Predicted Result}\\\hline\hline
True SC & Pred SC & AbsError \\\hline
441.1757 & 441.4429 & 0.2672 \\
408.0416 & 408.4984 & 0.4568 \\
380.4325 & 381.1115 &  0.6790\\
... & ... & ... \\
191.6724 & 191.5035 & 0.1689 \\
\hline\hline
\end{tabular}
}
\end{table}
\begin{table}
\caption{The MAE and MAPE of model NLR-SC, when tested on data of $N=40,45,50, 2\leq K \leq N$}
\label{table:Error-40-50}
\centerline{
\begin{tabular}{|c|c|c|c|} \hline
\multicolumn{4}{|c|}{Error Table}\\\hline\hline
       & MAE  & RMSE &MAPE \\ \hline
$N=40$ & 1.67  & 1.99 & 0.71\% \\
$N=45$ & 2.32 & 2.74 & 0.83\% \\
$N=50$ & 3.08 & 3.62 & 0.94\%\\ \hline
Average & 2.41 & 2.92 & 0.83\% \\
\hline\hline
\end{tabular}
}
\end{table}

Now we have 3 training sets, $Train_{N=40}$, $Train_{N=45}$, and $Train_{N=50}$. Each training set contains 5 training examples, and we divide the test data into 3 test sets by N, and refer to them as $Test_{N=40}$, $Test_{N=45}$, and $Test_{N=50}$. Table \ref{table:Sample-N-40} shows the data in $Train_{N=40}$ and $Test_{N=40}$. We train NLR-SC-N on $Train_{N=40}$ and test it on $Test_{N=40}$. We repeat this process for the other 2 training and test sets.  Prediction errors are calculated for each step and in total, as shown in Table \ref{table:Error-40-50}.
This example demonstrates how the two sub-models are combined, as well as the ability of NLR-SC to predict the SC for large N, using very little training data. MAPE is less than 1\%, which is much less than any model or algorithms that we have studied before. One of the reasons why this model performs very well is that it takes advantage of the theory of smoothed analysis. Because NLR-SC-K relies on the data patterns that relate to the worst-case and the average-case behaviour, it leads to very good prediction quality. Therefore NLR-SC-N, that is trained on the results of NLR-SC-K, also delivers good prediction results.

\subsection{Other Sorting Algorithms}

The biggest advantage of NLR-SC is that it can be applied to  algorithms other than Quicksort. In this work we also present experiments with NLR-SC for M3Quicksort, optimized Bubblesort and Mergesort. 

We can explain NLR-SC in three steps. First, it predicts the worst-case runtime of a given list length N. Then it predicts the average-case runtime of the same N. Finally, it predicts how the SC turns from worst-case runtime to the average-case runtime. Therefore, as long as we know the worst-case and average-case complexity of an algorithm and have some small amount of ground truth data, technically, the algorithm's SC can be predicted by NLR-SC.
Although the data shape of the SC of other algorithms is different to Quicksort's, NLR-SC can easily adjust to this difference. All we need to do for the transition is to select the correct $lm$ model to predict the worst-case and average-case complexity, which is dictated by the theory. For instance, unlike Quicksort, Mergesort's worst-case and average-case complexity are both $O(Nlog(N))$. Therefore, in NLR-SC-K, we simply replace the features of the $lm$ models to $N\times \log (N)$ and N, then the NLR-SC for predicting the SC of Mergesort can be used as is.
%
%
%

\section{Detailed Results of NLR-SC}

\subsection{QuickSort}

We first train and test NLR-SC on Quicksort. We work with three sets of data. The first set contains data of $10 \leq N \leq 100, 2 \leq K \leq N$ and N increases by 5. The second set contains data of $100 \leq N \leq 500, 2 \leq K \leq N$ and N increases by 100. The last set contains data of $600 \leq N \leq 3000, 2 \leq K \leq N$ and N increases by 300. 
We define $train_{a-b}$ as a training set of $a \leq N \leq b$, $test_{a-b}$ as a test set of $a \leq N \leq b$, and $nls\&lm_{a-b}$ as a NLR-SC trained on $train_{a-b}$. In addition, we define $nls\&lm_{a-b}^t$ as the NLR-SC with $t$ number of $lm$ models in NLR-SC-K, which means, the $nls$ in NLR-SC-N is trained on $t$ number of training examples. For all NLR-SC examples we discussed in previous sections, their $t$ value is 5, i.e., they have 5 $lm$ models in NLR-SC-K for $K=2,3,4,5,N$. In our experiments we found that the higher the $t$ values, the more accurate the prediction. These terms are also listed in Table \ref{table:Terms-M2}.
\begin{table}[htbp]
\caption{Description of terms used in NLR-SC.}
\label{table:Terms-M2}
\begin{center}
\begin{tabular}{|l|p{8cm}|} \hline\hline
Term  & Description \\ \hline\hline
$train_{a-b}$ & A training set of $a \leq N \leq b, 2\leq K \leq N$ \\\hline
$nls\&lm_{a-b}^t$ & A NLR-SC with t $lm$ models in NLR-SC-K, trained on $train_{a-b}$ \\\hline
$test_{a-b}$ & A test set of $a \leq N \leq b, 2\leq K \leq N$ 
\\\hline\hline
\end{tabular}
\end{center}
\end{table}

\begin{table}[htbp]
\caption{MAE and MAPE of the predicted results of NLR-SC for Quicksort.}
\label{table:Result-M2}
\centerline{
\begin{tabular}{|c|c|c|c|c|c|c|c|} \hline\hline
Model & Error &$test_{40-50}$ & $test_{90-100}$& $test_{300-300}$ &$test_{500-500}$& $test_{1.2k-1.2k}$& $test_{3k-3k}$ \\ \hline
\multirow{2}{*}{$nls\&lm_{10-20}^5$} & MAE & 2.41 & 13.93 & 116.20 & 257.05 & 864 &2646.65 \\\cline{2-8}
                                     & MAPE& 0.83\% & 1.68\% & 2.99\% & 3.44\% & 3.91\% &4.04\% \\\hline \hline                                     
\multirow{2}{*}{$nls\&lm_{10-80}^5$} & MAE & 2.73 & 16.65 & 143.38 & 321.49 & 1113.69 &3560.26 \\\cline{2-8}
                                     & MAPE& 0.96\% & 2.08\% & 3.83\% & 4.45\% & 5.18\% &5.50\% \\\hline\hline 
\multirow{2}{*}{$nls\&lm_{100-500}^5$} & MAE& 2.87 & 16.82 & 150.57 & 342.33 & 1209.75 &4941.15 \\\cline{2-8}
                                     & MAPE& 1.03\% & 2.10\% & 4.07\% & 4.82\% & 5.76\% &6.26\% \\\hline  \hline                                    
\multirow{2}{*}{$nls\&lm_{10-20}^6$} & MAE & 2.15 & 12.44 & 103.00 & 225.93 & 741.59 &2172.66 \\\cline{2-8}
                                     & MAPE& 0.75\% & 1.52\% & 2.70\% & 3.09\% & 3.48\% &3.52\% \\\hline\hline 
\multirow{2}{*}{$nls\&lm_{10-20}^7$} & MAE & 1.93 & 11.24 & 92.67 & 201.82 & 647.85 & 1879.58 \\\cline{2-8}
                                     & MAPE& 0.68\% & 1.39\% & 2.48\% & 2.83\% & 3.15\% &3.14\% \\\hline  \hline                                   
\multirow{2}{*}{$nls\&lm_{10-20}^8$} & MAE & 1.75 & 10.24 & 84.26 & 182.42 & 582.50 &1699.65 \\\cline{2-8}
                                     & MAPE& 0.62\% & 1.28\% & 2.29\% & 2.61\% & 2.89\% &2.86\% \\\hline\hline
\end{tabular}
}
\end{table}
Table \ref{table:Result-M2} shows the MAE and MAPE of NLR-SC trained on different training sets, tested on various test sets. The first model $nls\&lm_{10-20}^5$ is trained on data of $N=10,15,20$, with 5 $lm$ models in NLR-SC-K. The second model $nls\&lm_{10-80}^5$ is similar to the first one, but trained on bigger dataset of $10\leq N\leq 80$, and so on. The $nls\&lm_{10-20}^6$ model, on the other hand, has the same training set as the first model, but has 6 $lm$ models in NLR-SC-K. Based on the test results, we can see that simply increasing the size of the training data for NLR-SC, does not improve the accuracy, but increasing the number of $lm$ models in NLR-SC-K, which also means increasing the training examples of NLR-SC-N, improves the accuracy.

Comparing to Table \ref{table:Result-M1}, Table \ref{table:Result-M2} shows that NLR-SC has a much better prediction accuracy than TLR-SC. The average MAPE of NLR-SC on any test set is around 0.5\% to 3\%. Unlike TLR-SC, the error does not increase quickly when the size of the test set increases. Due to the computational requirements of modular smoothed analysis, the maximum test data we can collect through the modular approach is $N=3000$, and NLR-SC performs well on it, with around 4\% MAPE. Based on current test results, we expect NLR-SC to perform well on even larger $N$, but we currently lack the ground truth data to showcase this.

The most encouraging point of the test results is that we only use very few training examples to achieve such accurate prediction results. For model $nls\&lm_{10-20}^5$, the number of training examples it trained on is only 15. It means that if we apply NLR-SC to other algorithms, and only use the data collected through an experimental approach, we can still predict the SC of $N$ greater than 1,000. To our best knowledge, for now, only modular smoothed analysis can achieve that, and only for a restricted class of sorting algorithms (Quicksort and M3Quicksort) so far.

\subsection{M3Quicksort}
We also applied NLR-SC to other algorithms. Table \ref{table:Result-MOT-M2} shows MAE and MAPE of the prediction for M3Quicksort, for which we collected data through the modular approach. Note that the recurrence equation of modular smoothed analysis for M3Quicksort only works for $K \geq 4$, therefore, we do not have the SC of $K=2$, and $K=3$. And because of the factorial calculation in the formula, the maximum data we can collect is $N\leq 130$, $4\leq K \leq N$. Prediction on higher N values cannot be validated due to the lack of ground truth data.

NLR-SC relies on the data of $K=2$ and $K=3$ to ensure the accuracy of NLR-SC-K, as well as the whole model, but the modular approach formula of M3Quicksort cannot supply them. We have to make NLR-SC work on data of $K=4,5,6,7$ instead, and use $Train_{15-25}$ instead of $Train_{10-20}$, because when $N=10,K=7$, the SC is not at all the worst-case behavior, while when $N=15,K=7$, the SC is comparably closer to the worst-case behavior. This is why $nls\&lm_{40-50}^5$ gives a better result than $nls\&lm_{15-25}^5$: for $N=40,K=7$, the SC is closer to the worst-case behaviour than for $N=15,K=7$. Besides, similar to the Quicksort results, adding in more $lm$ models in NLR-SC-K (i.e. more training examples in NLR-SC-N) can improve the accuracy.
\begin{table}[htbp]
\caption{MAE and MAPE of the predicted results of NLR-SC on M3Quicksort.}
\label{table:Result-MOT-M2}
\begin{center}
\begin{tabular}{|c|c|c|c|} \hline\hline
 Model&Error &$test_{40-50}$ & $test_{90-100}$ \\ \hline
\multirow{2}{*}{$nls\&lm_{15-25}^5$}&MAE & 1.69 & 12.35  \\\cline{2-4}
                                    &MAPE & 0.72\% & 1.82\%  \\\hline\hline
\multirow{2}{*}{$nls\&lm_{15-25}^7$}&MAE & 1.52 & 12.36  \\\cline{2-4}
                                    &MAPE & 0.64\% & 1.82\%  \\\hline\hline
\multirow{2}{*}{$nls\&lm_{40-50}^5$}&MAE & 1.39 & 9.06  \\\cline{2-4}
                                    &MAPE & 0.61\% & 1.45\%  \\\hline\hline
\multirow{2}{*}{$nls\&lm_{40-50}^7$}&MAE & 1.17 & 8.37  \\\cline{2-4}
                                    &MAPE & 0.52\% & 1.34\%  \\\hline\hline
\end{tabular}
\end{center}
\end{table}

\subsection{Optimized Bubblesort}

We also test NLR-SC on optimized Bubblesort, for which we collected data through the experimental approach. The size of the dataset is 50, with $3\leq N\leq 15$ and $2\leq K\leq N$. Due to the limited computation power, the higher the N value, the less data we can collect, e.g., for $N=15$ we can only collect ground truth data up to $K=4$.

The original Bubblesort's worst-case runtime is equal to its  average-case runtime. The optimized Bubblesort that we used in this paper has better performance on part of the input lists, but its average-case complexity remains $O(N^2)$, which is different from Quicksort. The good news is that NLR-SC can easily adjust to such difference. By simply replacing the feature $N\times \log(N)$ in $lm$ models to $N^2$, NLR-SC is  able to capture the SC of algorithms whose average-case and worst-case complexity are both $O(N^2)$.
Considering the limited amount of the ground truth data, we reduce the number of $lm$ models as well as the number of parameters in $nls$ to 4. Thus, NLR-SC only uses 12 training examples, that follow either worst-case or average-case behavior.

Table \ref{table:Result-BS-M2} shows the error of the predicted results. Due to the shortage of ground truth data, we only test NLR-SC on a test set of $N=9$ and $N=10$. The predicted results are fairly good, with MAE of 0.05 and MAPE of 0.16\%. 
\begin{table}[htbp]
\caption{MAE and MAPE of the predicted results of NLR-SC on optimized Bubblesort.}
\label{table:Result-BS-M2}
\begin{center}
\begin{tabular}{|c|c|c|c|} \hline\hline
Model & Error &$Test_{9-9}$ & $Test_{10-10}$   \\ \hline
\multirow{2}{*}{$nls\&lm_{5-8}^4$} & MAE & 0.05 & 0.07   \\ \cline{2-4}
                                   & MAPE& 0.16\% & 0.17\%  \\ \hline\hline
\end{tabular}
\end{center}
\end{table}

\begin{figure}[pt]
\centering
\subfloat[Subfigure 1 list of figures text][Bubblesort]{
\includegraphics[width=0.5\textwidth]{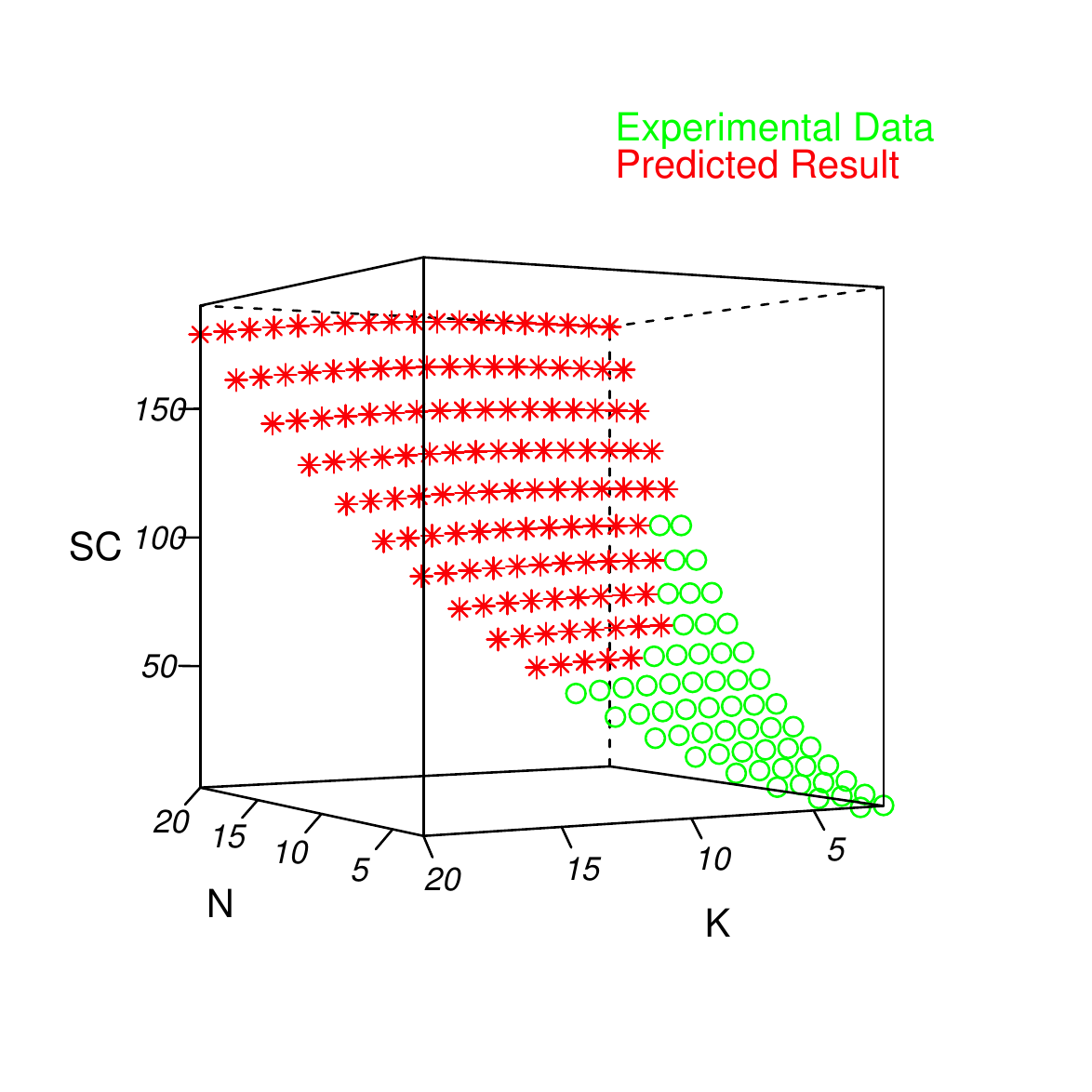}
\label{M2-BS-Cloud}}
\subfloat[Subfigure 2 list of figures text][Mergesort]{
\includegraphics[width=0.5\textwidth]{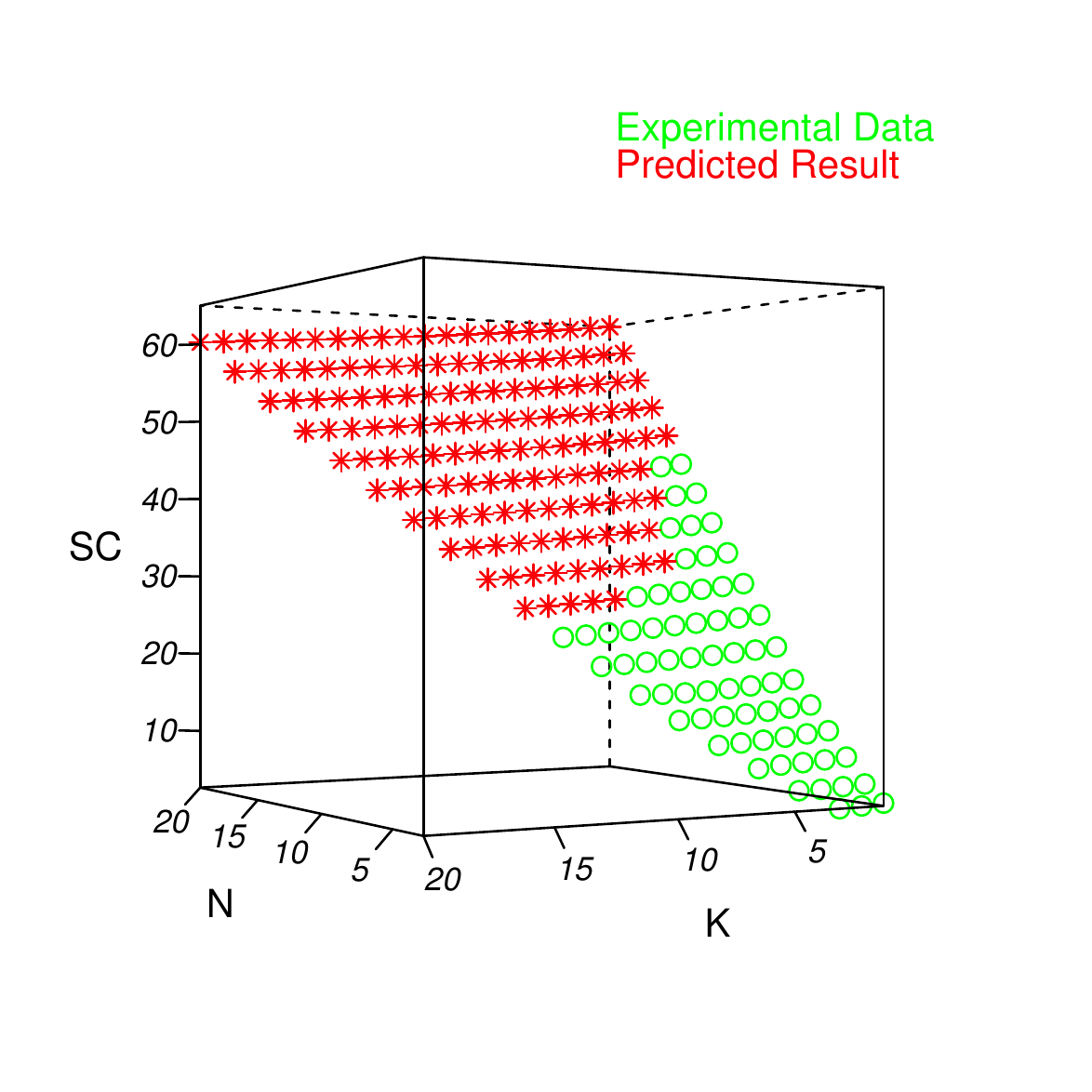}
\label{M2-MS-Cloud}}
\caption{Plot of the predicted results of NLR-SC on Bubblesort and Mergesort. The model is trained on experimental data, which are marked as green spots.}
\label{M2-BSMS-Cloud}
\end{figure}
Figure \ref{M2-BS-Cloud} shows the plot of the predicted results. The green spots are the experimentally collected data, while red stars are predicted results of NLR-SC, where the real values of these data are unknown and cannot be collected through an experimental approach. We observe from the plot that the predicted results seem to agree with the trend of the experimental data, but have no ground truth data to numerically validate them. 

\subsection{Mergesort}

\begin{figure}[h!]
\centering
\subfloat[Subfigure 1 list of figures text][Quicksort]{
\includegraphics[width=0.5\textwidth]{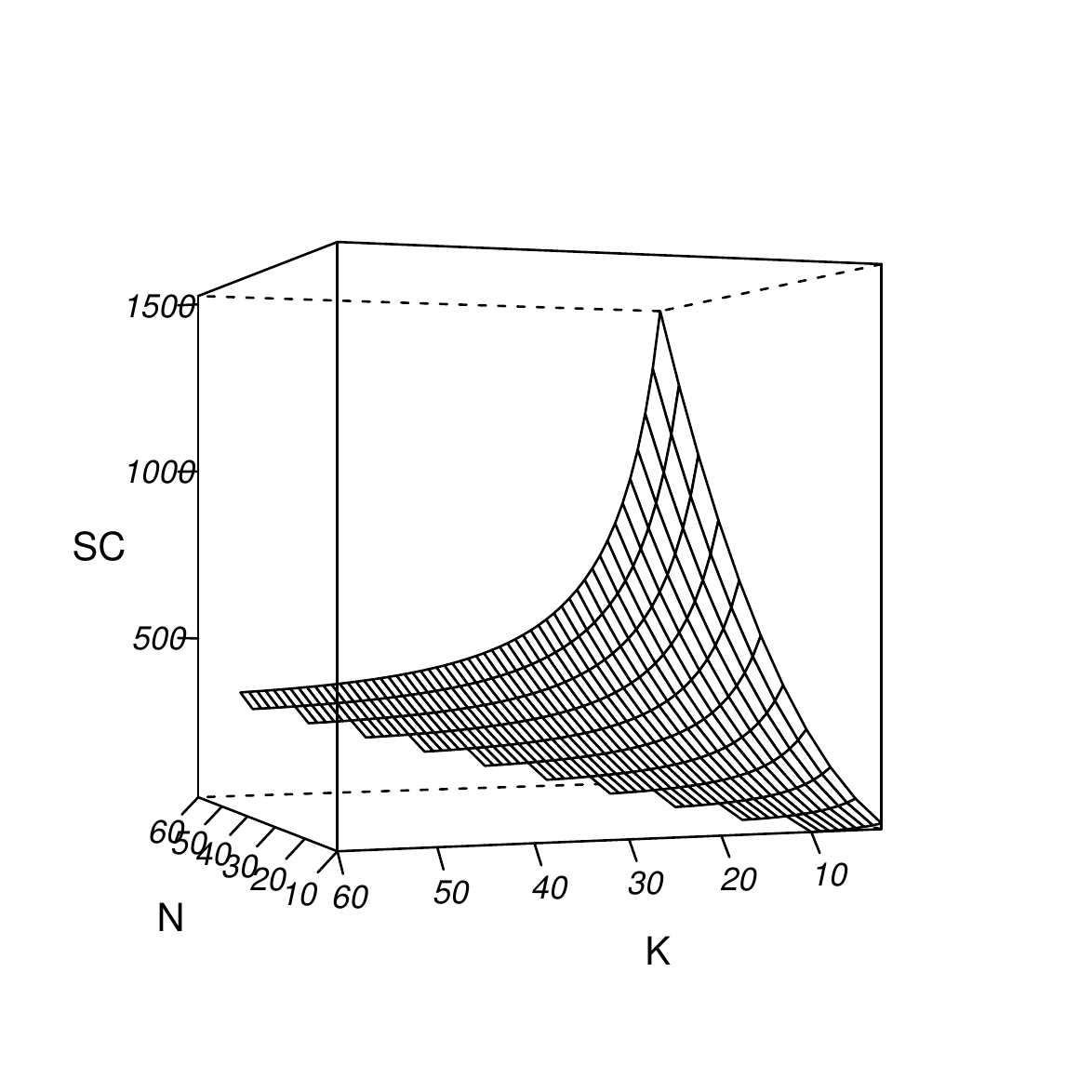}
\label{QS-60}}
\subfloat[Subfigure 2 list of figures text][M3Quicksort]{
\includegraphics[width=0.5\textwidth]{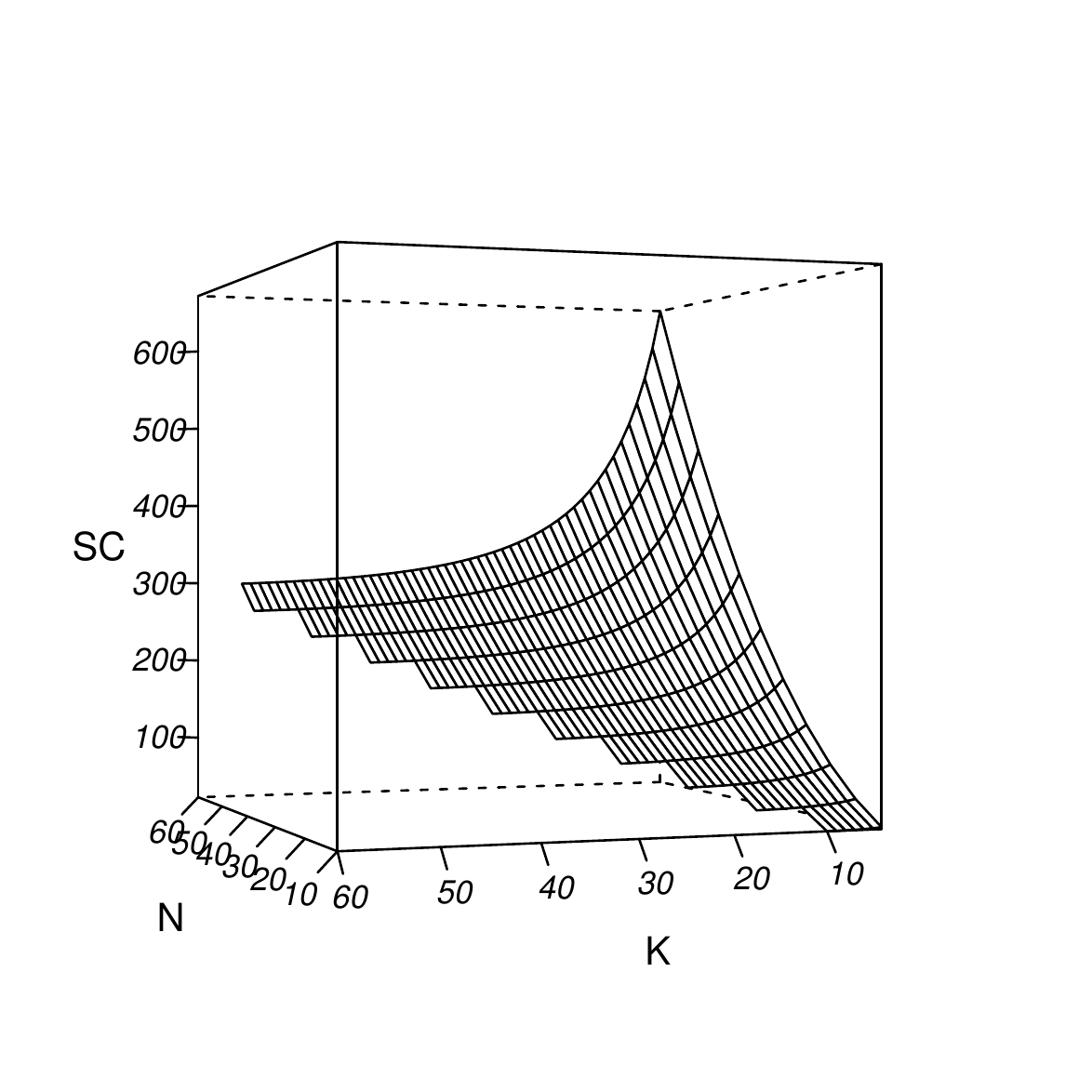}
\label{MOT-60}}
\\
\subfloat[Subfigure 3 list of figures text][Bubblesort]{
\includegraphics[width=0.5\textwidth]{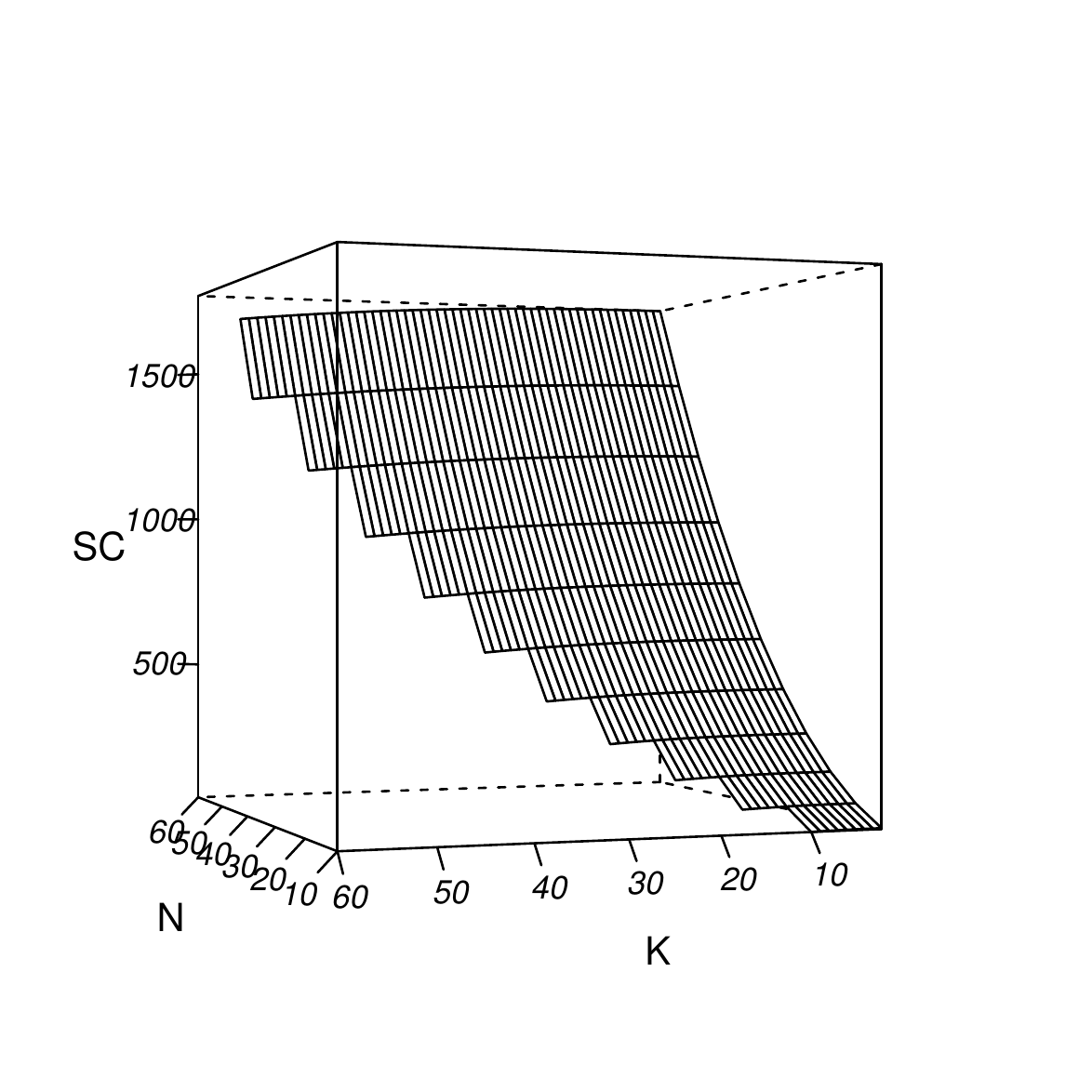}
\label{BS-60}}
\subfloat[Subfigure 4 list of figures text][Mergesort]{
\includegraphics[width=0.5\textwidth]{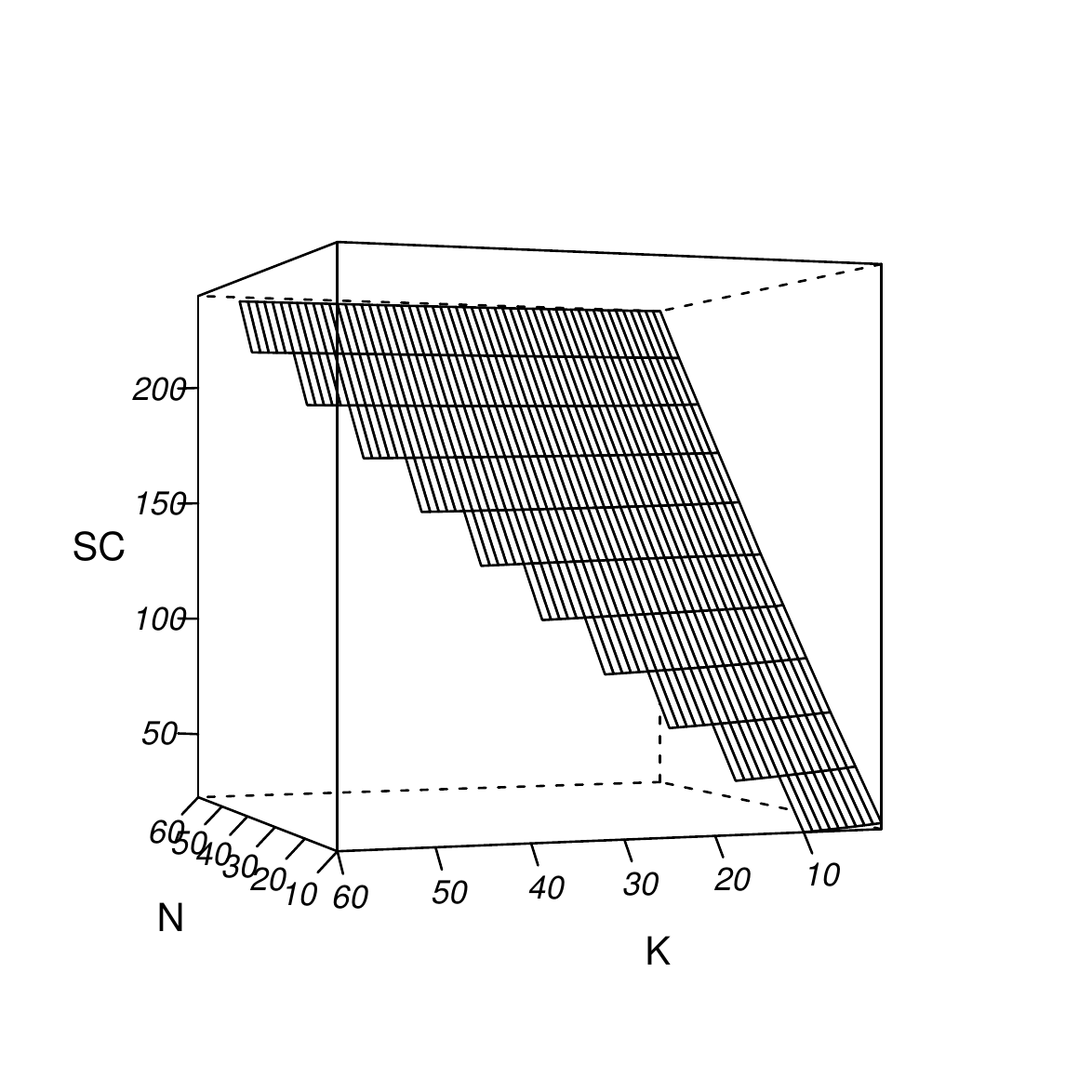}
\label{MS-60}}
\caption{Plot of the predicted SC of 4 sorting algorithms, for $N\leq 60$,$2\leq K\leq N$.}
\label{All-60}
\end{figure}

Similar to Bubblesort, the data of Mergesort is collected through the experimental approach. The total number of examples is 50, with data of $3\leq N\leq 15$ and $2\leq K\leq N$.
Both the worst-case complexity and the average-case complexity of Mergesort is $O(Nlog(N))$, therefore, we change all $N^2$ features in $lm$ models to $N\times log(N)$. Similar to Bubblesort, because of the limited ground truth data, we reduce the number of $lm$ models as well as the number of parameters in $nls$ to 4. 
Tested on $N=9$ and $N=10$, the error of the predicted results of NLR-SC for Mergesort are shown in Table \ref{table:Result-BS-M2}. Figure \ref{M2-MS-Cloud} shows the plot of the predicted results. Although the data shape of Mergesort is different to Bubblesort, the SC prediction accuracy is still high.
\begin{table}[h!]
\caption{MAE and MAPE of the predicted results of NLR-SC on Mergesort.}
\label{table:Result-MS-M2}
\centerline{
\begin{tabular}{|c|c|c|c|} \hline\hline
Model & Error &$Test_{9-9}$ & $Test_{10-10}$   \\ \hline
\multirow{2}{*}{$nls\&lm_{5-8}^4$} & MAE & 0.05 & 0.05   \\ \cline{2-4}
                                   & MAPE& 0.20\% & 0.22\%  \\ \hline\hline
\end{tabular}
}
\end{table}
Figure \ref{All-60} shows the predicted SC of Quicksort, M3Quicksort, optimized Bubblesort and Mergesort given by NLR-SC, for $N\leq 60$, $2\leq K\leq N$. 

\section{Conclusion}
In this work, we present two regression models for predicting the Smoothed Complexity (denoted as SC) of sorting algorithms.
The first model, TLR-SC, uses linear regression to predict the complex data surface of the SC. It transfers a nonlinear relationship into a linear one, by transforming the original features. TLR-SC has a simple structure and is very efficient, although it is aimed at Quicksort and is not easily transferable to other sorting algorithms.
The second model, NLR-SC, takes an iterative approach. It divides the surface fitting problem into multiple curve fitting problems and by predicting curves one by one, it gradually predicts the entire surface. NLR-SC takes advantage of the theory of the smoothed analysis, which leads to its very good performance. With 15 training examples, NLR-SC can predict the SC of Quicksort for $N \geq 3000$, with around 3\% Mean Absolute Percentage Error. The rules NLR-SC relies on are general and grounded on the theory of SC: when the list perturbation parameter $K$ approaches 1, the SC of any sorting algorithm follows the worst-case behavior, and when K approaches N, the SC follows the average-case behavior. Therefore, technically, as long as we know the worst-case and average-case complexity of a sorting algorithm, and have a small amount of training examples, we can use NLR-SC to predict the SC of
the algorithm. 

In this work we study the SC of four sorting algorithms and show that our prediction models deliver high quality results.
A large part of our study focuses on data collection and analysis, since to start with, there is very limited availability of ground truth data, to test our prediction models. By getting a good understanding of the data shape and the relationship between features and the SC, we construct two predictive models that work very well with limited training data, and show that the shortage of data can be made up by the adequate background knowledge, in our case provided by the theory of SC. Our work fills the gap between known theoretical and empirical results on the behaviour of the Smoothed Complexity of sorting algorithms. By taking a machine learning approach, we build predictive models that are scalable for large input sizes, therefore advancing currently existing techniques.
In the future, we plan to further analyze the potential of our prediction models, by studying other (than sorting) type of algorithms working on discrete data. 

\section{Acknowledgements}

This work was supported by Science Foundation Ireland under grant 07/CE/I1147 and 07/IN.1/I977.

\bibliographystyle{natbib}

\bibliography{predicting-sc} 

\begin{thebibliography}{}

\bibitem[Arthur {\em et~al.}(2009)Arthur, Manthey, and Roglin]{arthur2009k}
Arthur, D., Manthey, B., and Roglin, H. (2009).
\newblock k-means has polynomial smoothed complexity.
\newblock In {\em Foundations of Computer Science, 2009. FOCS'09. 50th Annual
  IEEE Symposium on\/}, pages 405--414. IEEE.

\bibitem[Banderier {\em et~al.}(2003)Banderier, Beier, and
  Mehlhorn]{banderier2003smoothed}
Banderier, C., Beier, R., and Mehlhorn, K. (2003).
\newblock Smoothed analysis of three combinatorial problems.
\newblock In {\em Mathematical Foundations of Computer Science 2003\/}, pages
  198--207. Springer.

\bibitem[Blum and Dunagan(2002)Blum and Dunagan]{Blum:2002:SAP:545381.545499}
Blum, A. and Dunagan, J. (2002).
\newblock Smoothed analysis of the perceptron algorithm for linear programming.
\newblock In {\em Proceedings of the thirteenth annual ACM-SIAM symposium on
  Discrete algorithms\/}, SODA '02, pages 905--914, Philadelphia, PA, USA.
  Society for Industrial and Applied Mathematics.

\bibitem[Deshpande and Spielman(2005)Deshpande and
  Spielman]{deshpande2005improved}
Deshpande, A. and Spielman, D.~A. (2005).
\newblock Improved smoothed analysis of the shadow vertex simplex method.
\newblock In {\em Foundations of Computer Science, 2005. FOCS 2005. 46th Annual
  IEEE Symposium on\/}, pages 349--356. IEEE.

\bibitem[Hastie {\em et~al.}(2013)Hastie, Tibshirani, and
  Friedman]{hastie:slbook13}
Hastie, T., Tibshirani, R., and Friedman, J. (2013).
\newblock {\em The Elements of Statistical Learning\/}.
\newblock Springer Series in Statistics.

\bibitem[Hennessy and Schellekens(2014)Hennessy and
  Schellekens]{Schellekens2013ModularM3Q}
Hennessy, A. and Schellekens, M. (2014).
\newblock Modular smoothed analysis of median-of-three quicksort.
\newblock {\em Technical Report, University College Cork, Ireland. Submitted to
  Discrete Mathematics\/}.

\bibitem[Knuth(1973)Knuth]{DBLP:books/aw/Knuth73}
Knuth, D.~E. (1973).
\newblock {\em The Art of Computer Programming, Volume III: Sorting and
  Searching\/}.
\newblock Addison-Wesley.

\bibitem[Motulsky(2004)Motulsky]{motulsky2004fitting}
Motulsky, H. (2004).
\newblock {\em Fitting models to biological data using linear and nonlinear
  regression: a practical guide to curve fitting\/}.
\newblock OUP USA.

\bibitem[Schellekens(2008)Schellekens]{schellekens2008modular}
Schellekens, M. (2008).
\newblock {\em A Modular Calculus for the Average Cost of Data Structuring:
  Efficiency-Oriented Programming in MOQA\/}.
\newblock Springer.

\bibitem[Schellekens {\em et~al.}(2014)Schellekens, Hennessy, and
  Shi]{Schellekens2013Modular}
Schellekens, M., Hennessy, A., and Shi, B. (2014).
\newblock Modular smoothed analysis.
\newblock {\em Technical Report, University College Cork, Ireland. Submitted to
  Discrete Mathematics\/}.

\bibitem[Shi(2013)Shi]{claire:mst}
Shi, B. (2013).
\newblock {\em A Machine Learning Approach For Estimating The Smoothed
  Complexity Of Sorting Algorithms\/}.
\newblock Master's thesis, University College Cork, Cork, Ireland.

\bibitem[Spielman and Teng(2001)Spielman and Teng]{spielman2001smoothed}
Spielman, D. and Teng, S.-H. (2001).
\newblock Smoothed analysis of algorithms: Why the simplex algorithm usually
  takes polynomial time.
\newblock In {\em Proceedings of the thirty-third annual ACM symposium on
  Theory of computing\/}, pages 296--305. ACM.

\bibitem[Spielman and Teng(2006)Spielman and Teng]{spielman2006smoothed}
Spielman, D.~A. and Teng, S. (2006).
\newblock Smoothed analysis of algorithms and heuristics.
\newblock {\em LONDON MATHEMATICAL SOCIETY LECTURE NOTE SERIES\/}, {\bf 331},
  274.

\bibitem[{Spielman} and {Teng}(2002){Spielman} and
  {Teng}]{spielman2002smoothed}
{Spielman}, D.~A. and {Teng}, S.-H. (2002).
\newblock {Smoothed analysis of algorithms}.
\newblock {\em ArXiv Mathematics e-prints\/}.

\bibitem[Spielman and Teng(2009)Spielman and Teng]{spielman2009smoothed}
Spielman, D.~A. and Teng, S.-H. (2009).
\newblock Smoothed analysis: an attempt to explain the behavior of algorithms
  in practice.
\newblock {\em Communications of the ACM\/}, {\bf 52}(10), 76--84.

\bibitem[Witten {\em et~al.}(2011)Witten, Frank, and Hall]{witten2011data}
Witten, I.~H., Frank, E., and Hall, M.~A. (2011).
\newblock {\em Data Mining: Practical Machine Learning Tools and Techniques\/}.
\newblock Morgan Kaufmann.

\end{thebibliography}

\end{document}